%% file: main.tex
\documentclass[11pt,fceqn]{article}

\usepackage{fullpage}
\usepackage{algorithm}
\usepackage{algpseudocode}
\usepackage{amsmath,amsthm,amsfonts}
\usepackage{amsmath}

\usepackage{amssymb}
\usepackage{color}
\usepackage{mathrsfs} 
\usepackage{enumitem}
\usepackage{bm}
\usepackage{multirow}
\usepackage{booktabs}
\usepackage{makecell}
\usepackage{graphicx}
\usepackage{subcaption}
\usepackage{comment}
\usepackage{cases}
\usepackage{appendix}
\usepackage{tikz}
\usetikzlibrary{arrows,shapes}
\usepackage{marginnote}
\usepackage{tcolorbox}

\usepackage[colorlinks, linkcolor=blue, citecolor=blue, urlcolor=red]{hyperref}

\include{notations}

\numberwithin{equation}{section}

\theoremstyle{definition}
\newtheorem{mainthm}{Theorem}
\newtheorem{thm}{Theorem}[section]
\newtheorem{lem}[thm]{Lemma}
\newtheorem{coro}[thm]{Corollary}

\newtheorem{prop}[thm]{Proposition}
\newtheorem{defn}{Definition}[section]
\newtheorem{assumption}[thm]{Assumption}

\newcommand\dd{\mathrm{d}}

\usepackage{cleveref}
\usepackage{tikz}
\usetikzlibrary{shapes.geometric, arrows, positioning}

\tikzset{
   mybox/.style  = {draw, rectangle, minimum width=3cm, minimum height=0.8cm, text centered, text width=2.4cm,   
  font=\normalsize},
  box/.style  = {draw, rectangle, minimum width=3cm, minimum height=0.8cm, text centered, text width=3.6cm,   
  font=\normalsize},
   myarrow/.style = {line width=0.2pt, draw=black, -triangle 60, postaction={draw, line width=0.2pt, shorten >=10pt,-}}
}

\tikzstyle{arrow} = [->, >=stealth, -triangle 60]

\allowdisplaybreaks

\DeclareMathOperator{\E}{\mathbb{E}}
\makeatletter
\newcommand{\leqnomode}{\tagsleft@true}
\newcommand{\reqnomode}{\tagsleft@false}
\makeatother

\begin{document}

\title{On the Hyperparameters in Stochastic Gradient Descent with Momentum}

\author{Bin Shi\thanks{State Key Laboratory of Scientific and Engineering Computing, Academy of Mathematics and Systems Science, Chinese Academy of Sciences, Beijing, China. Email: \textbf{shibin@lsec.cc.ac.cn} } 
}

\maketitle
\begin{abstract}

Following the same routine as~\cite{shi2020learning}, we continue to present the theoretical analysis for \textit{stochastic gradient descent with momentum} (SGD with momentum) in this paper. Differently, for SGD with momentum, we demonstrate that the two hyperparameters together, the learning rate and the momentum coefficient, play a significant role in the linear convergence rate in non-convex optimizations. Our analysis is based on using a~\textit{hyperparameters-dependent stochastic differential equation} (hp-dependent SDE) that serves as a continuous surrogate for SGD with momentum. Similarly, we establish the linear convergence for the continuous-time formulation of SGD with momentum and obtain an explicit expression for the optimal linear rate by analyzing the spectrum of the Kramers-Fokker-Planck operator. By comparison, we demonstrate how the optimal linear rate of convergence and the final gap for SGD only about the learning rate varies with the momentum coefficient increasing from zero to one when the momentum is introduced. Then, we propose a mathematical interpretation of why, in practice, SGD with momentum converges faster and is more robust in the learning rate than standard stochastic gradient descent (SGD). Finally, we show the Nesterov momentum under the presence of noise has no essential difference from the traditional momentum.

%
\vspace{.1in}

\noindent {\bf Keywords:} nonconvex optimization, stochastic gradient descent with momentum, hp-dependent SDE, hp-dependent kinetic Fokker-Planck equation, Kramers operator, Kramers-Fokker-Planck operator, Nesterov momentum

\vspace{.07in}

\noindent {\bf AMS 2000 subject classification:} 35K10,~35P05,~65C30,~68R01,~90C26,~90C30 

\end{abstract}
\thispagestyle{empty}
\setcounter{page}{0}


\input{intro}

\input{prelim}

\input{result}

\input{qualitative}

\input{quantitative}

\input{nesterov}
\input{conclu}

{\small
\subsection*{Acknowledgments}
We would like to thank Yu Sun help us practically run deep learning at the University of California, Berkeley.


\bibliographystyle{alpha}
\bibliography{sigproc}
}

\newpage
\appendix

\input{appendix/proof_kin_sde}

\input{appendix/discrete}

\input{appendix/Fokker-Planck}

\end{document}

%% file: intro.tex
\section{Introduction}
\label{sec: intro}

For a long time, it has been considered a core and fundamental topic to build theories for optimization algorithms, which leads, in practice, to design new algorithms for accelerating and improving performance. Recently, with the blossoming of machine learning, people have spotlighted gradient-based algorithms. However, the mechanism behind them is still mysterious and undiscovered. Significantly, the non-convex structure brings about new and urgent challenges for modern theoreticians.  

Recall the minimization problem of a non-convex function $f$ is defined in terms of an expectation:  
\[
f(x) = \mathbb{E}_{\zeta}[f(x; \zeta)],
\]
where the expectation is over the randomness embodied in $\zeta$. Empirical risk minimization, averaged over $n$ data points, is a simple and special case, shown as
\[
f(x) = \frac{1}{n} \sum_{i=1}^{n} f_i(x),
\]
where $x$ denotes a parameter and the datapoint-specific loss, $f_i(x)$, is indexed by $i$. When $n$ is large, it is prohibitively expensive to compute the full gradient of the objective function. Hence, the algorithms for gradients with incomplete information (noisy gradient) are adopted widely in practice.

\begin{figure}[htpb!]
\centering
\begin{minipage}[t]{0.45\linewidth}
\centering
\includegraphics[scale=0.14]{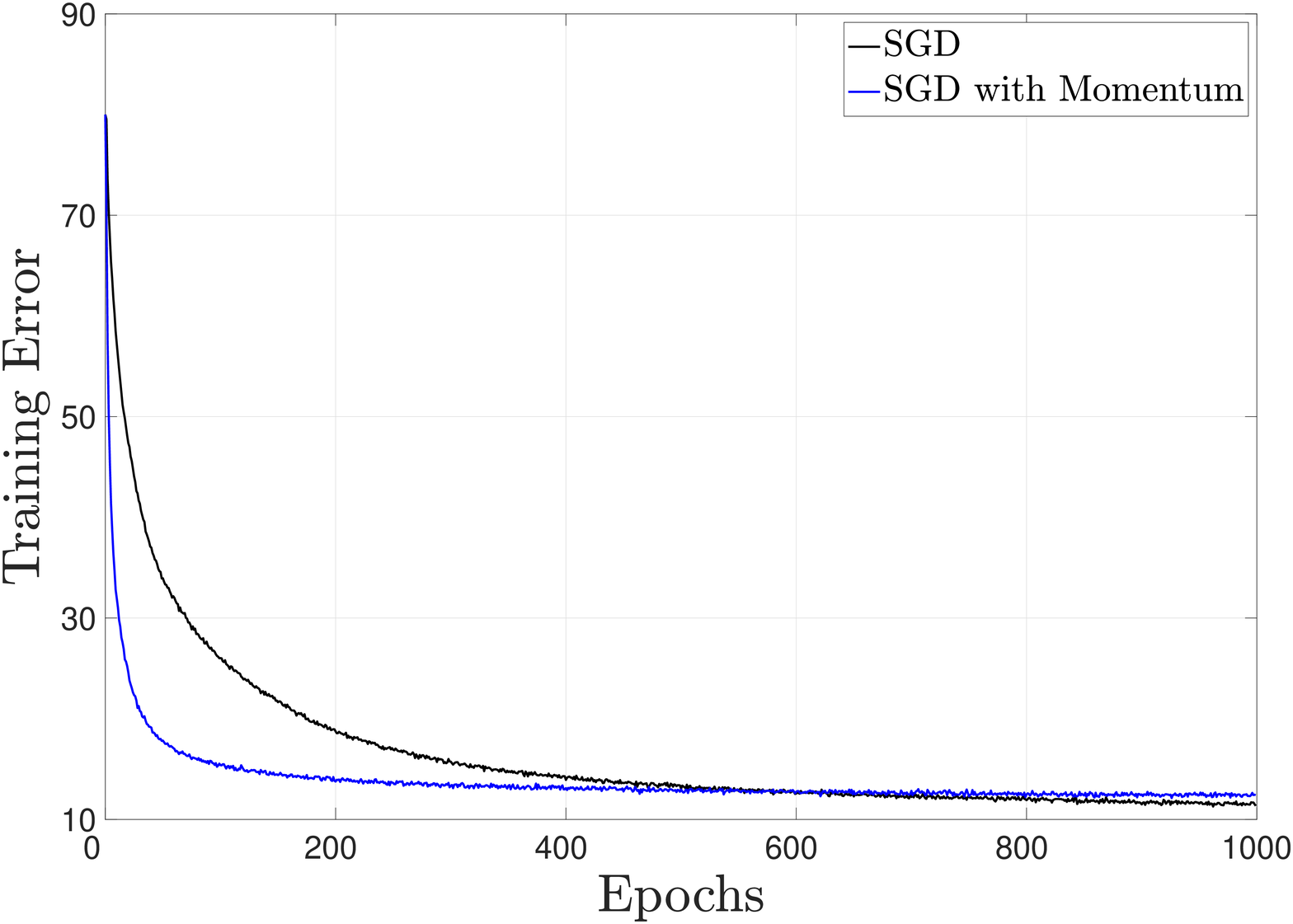}
\end{minipage}
\begin{minipage}[t]{0.45\linewidth}
\centering
\includegraphics[scale=0.14]{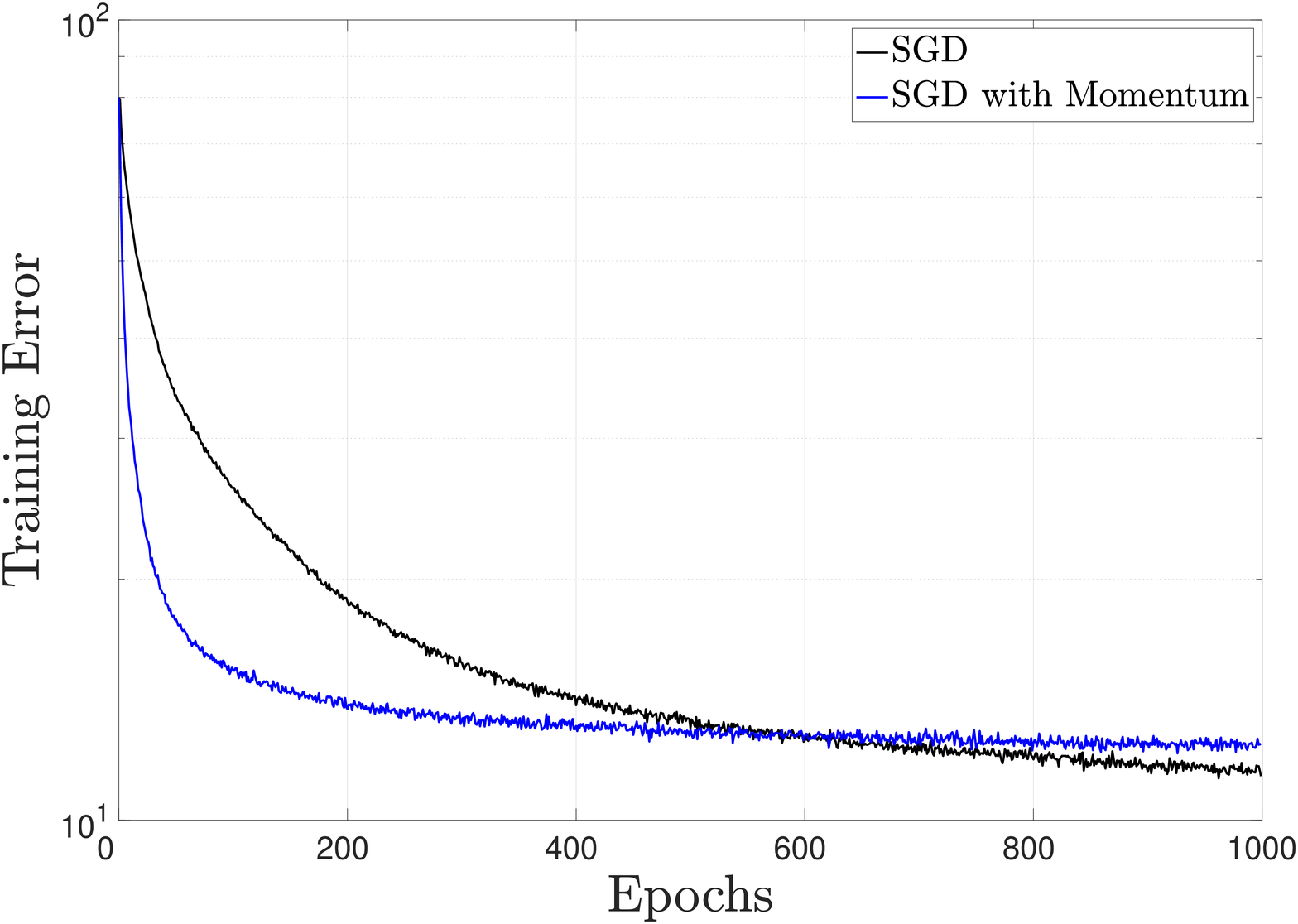}
\end{minipage}
\caption{\small The comparison for the training error between SGD and SGD with momentum. The setting is a 20-layer convolutional neural network on CIFAR-10~\cite{krizhevsky2009learning} with a mini-batch size of 128.~\textbf{Learning Rate:} $s = 0.01$. \textbf{Momentum Coefficient:} $\alpha = 0.9$. \textbf{Left:} Standard Scale; \textbf{Right:} Logarithmic Scale. See~\cite{he2016deep} for further investigation of this phenomenon.} 
\label{fig: msgd-deep}
\end{figure}

Recall standard~\textit{stochastic gradient descent}, shortened to SGD,
\begin{equation*}
\label{eqn: sgd}
x_{k+1} = x_{k} - s \nabla f(x_{k}) + s \xi_{k},
\end{equation*}
with any initial $x_0 \in \mathbb{R}^d$, where $\xi_k$ denotes the noise at the $k^{th}$ iteration. 
In this paper, we consider the most popular variant of SGD --- SGD~\textit{with momentum} 
\begin{equation*}
\label{eqn: sgd+momentum}
x_{k+1} = x_{k} - s \nabla f(x_{k}) + s \xi_{k} + \alpha (x_{k} - x_{k-1})
\end{equation*}
with any initial $x_0, x_1 \in \mathbb{R}^d$, where $\alpha (x_{k} - x_{k-1})$ is named as momentum excluded in SGD. SGD with momentum has been practically proven to be the most effective setting and is widely adopted in deep learning~\cite{he2016deep}. An experimental comparison between SGD and SGD with momentum is shown in~\Cref{fig: msgd-deep}.

%

In Figure~\ref{fig: msgd-deep}, we compare SGD and SGD with momentum using the same learning rate. For the stable (final) training error  (the plateau part in the curve of the training error), SGD with momentum is a little larger than SGD; however, for the least epochs arriving at stabilization,  SGD with momentum is far less than SGD. Generally speaking, the momentum coefficient in SGD with momentum is usually set to $\alpha = 0.9$ or beyond in practice. Straightforwardly, people will ask: 

\vspace{0.2cm}
\begin{center}
\begin{tcolorbox}
\textbf{Why does SGD with momentum under the setting of momentum coefficient $\alpha = 0.9$ and beyond often perform well?} 
\end{tcolorbox}
\end{center}
\vspace{0.1cm}

\noindent Actually, it is still a mystery. Now, we can delve into it with two formal and academic questions: 

\begin{itemize}
\item  What happens to the iterative behavior of SGD when the momentum is introduced? Specifically,  how does the iterative behavior of SGD with momentum vary with the momentum coefficient $\alpha$ from $0$ to $1$ when the learning rate $s$ is fixed?
\item  How does the iterative behavior of SGD with momentum vary with the learning rate $s$ when the momentum coefficient $\alpha$ is fixed? 

\end{itemize} 
In other words, do we need to investigate the relationship between SGD and SGD with momentum? Where are the similarities? Where are the differences? In this paper, we will try to answer the questions above from the continuous surrogate. 
Before discussing the property of SGD with momentum about the hyperparameters, the learning rate $s$ and the momentum coefficient $\alpha$, we introduce, for convenience, a new mixing parameter  as
\[
\mu = \frac{(1 - \alpha)^2}{(1 + \alpha)^2} \cdot \frac{1}{s} \in \left( 0,\; \frac{1}{s} \right) \quad \text{such that} \quad \alpha = \frac{1 - \sqrt{\mu s}}{1 + \sqrt{\mu s}} \in (0, 1).
\]

\subsection{Continuous-time approximation}            
The continuous-time approximation, used widely to study the gradient-based optimization algorithms, is a newly-sprung and vital to assist in the proposal of fundamental new insights and understandings for the performance~\cite{su2016differential}. Due to its conceptual simplicity in the continuous setting,  many properties related to them have been discovered and developed~\cite{shi2018understanding}. Moreover, modern and advanced mathematics and physics methods can propose profound and powerful analyses for them~\cite{shi2020learning}.

Now, we construct the continuous surrogate for SGD with momentum as follows. Taking a small but nonzero learning rate $s$, let $t_{k} = k\sqrt{s}$ $(k = 0, 1, 2, \ldots)$ denote a time step and define $x_{k} = X_{s, \alpha}(t_{k})$ for some sufficiently smooth curve $X_{s, \alpha}(t)$.\footnote{The subscripts outline $X_{s, \alpha}(t)$ is dependent on learning rate $s$ and momentum coefficient $\alpha$.} Applying a Taylor expansion in the power of $\sqrt{s}$, we obtain:
\begin{equation}
\label{eqn: taylor-3}
\left\{
\begin{aligned}
& x_{k+1} = X_{s,\alpha}(t_{k+1}) =  x_k + \dot{X}_{s,\alpha}(t_{k}) \sqrt{s} + \frac{1}{2} \ddot{X}_{s,\alpha}(t_{k}) s + \frac{1}{6} \dddot{X}_{s,\alpha}(t_{k}) (\sqrt{s})^3 + O(s^{2}) \\
& x_{k-1}  = X_{s,\alpha}(t_{k-1})  =  x_k - \dot{X}_{s,\alpha}(t_{k}) \sqrt{s} + \frac{1}{2} \ddot{X}_{s,\alpha}(t_{k}) s - \frac{1}{6} \dddot{X}_{s,\alpha}(t_{k}) (\sqrt{s})^3 + O(s^{2})
\end{aligned}
\right.
\end{equation}
Let $W$ be a standard Brownian motion and, for the time being, assume that the noise term $\xi_k$ approximately follows the normal distribution with unit variance. Informally, this leads to
\begin{equation}
\label{eqn: noise-continuous}
\sqrt[4]{s} \xi_k = W(t_{k+1}) - W(t_{k}) = \sqrt{s} \dot{W}(t_k) + O(s).\footnote{Although a Brownian motion is not differentiable, the notation $\dot{W}$ has been used in~\cite{evans2012introduction, villani2006hypocoercive}.}
\end{equation}
Taking the straightforward transform for the SGD with momentum, we have
\[
\frac{x_{k+1} + x_{k-1} - 2x_k}{s} + \frac{2\sqrt{\mu}}{1 + \sqrt{\mu s}} \cdot \frac{x_{k} - x_{k-1}}{\sqrt{s}} + \nabla f(x_{k})  = \xi_k. 
\]
Then, we plug the previous two displays,~\eqref{eqn: taylor-3} and~\eqref{eqn: noise-continuous}, into SGD with momentum and get
\[
\ddot{X}_{s, \alpha}(t_{k}) + O(s) + 2\sqrt{\mu} \dot{X}_{s, \alpha}(t_k) + O(\sqrt{s}) + \nabla f(x_{k}) = \sqrt[4]{s} \dot{W}(t_k)  + O(s^{\frac{3}{4}}).
\] 
Considering the $O(\sqrt[4]{s})$-approximation, that is, retaining both $O(1)$ and $O(\sqrt[4]{s})$ terms but ignoring the smaller terms, we obtain a \textit{hyperparameter-dependent stochastic differential equation} (hp-dependent SDE) 
\begin{equation}
\label{eqn: lr-momentum-sde}
\ddot{X}_{s, \alpha} + 2\sqrt{\mu} \dot{X}_{s, \alpha}  + \nabla f(X_{s, \alpha}) = \sqrt[4]{s} \dot{W},  
\end{equation}
where the initial condition is the same value $X_{s, \alpha}(0) = x_0$ and $\dot{X}_{s, \alpha}(0)  = (x_1 - x_0)/\sqrt{s}$. 
Here, we write down the standard form of the hp-dependent SDE~\eqref{eqn: lr-momentum-sde} as
\begin{equation}
\label{eqn: lr-momentum-sde-standard}
\left\{ \begin{aligned}
          & dX_{s, \alpha} = V_{s, \alpha} , \\
          & dV_{s, \alpha} =  \left(- 2\sqrt{\mu} X_{s, \alpha} - \nabla f(X_{s, \alpha}) \right) dt + \sqrt[4]{s} dW.
         \end{aligned} \right.
\end{equation}

\subsection{An intuitive analysis}
\label{subsec: intuitive}

In~\cite{shi2020learning}, the authors derive the continuous surrogate for SGD,~\textit{lr-dependent} SDE, 
\begin{align}
\label{eqn: lr-sde}
d X_s= - \nabla f(X_s) d t +  \sqrt{s} d W,
\end{align}
where the initial is the same value $x_0$ as its discrete counterpart and $W$ is the standard Brownian motion.
By the corresponding lr-dependent Fokker-Planck equation
\begin{equation}
\label{eqn: lr-dep-FP}
\frac{\partial \rho_s}{\partial t} = \nabla \cdot \left( \rho_s \nabla f \right) + \frac{s}{2} \Delta \rho_s,
\end{equation}
 they show the expected excess risk evolves with time as
\[ \mathbb{E}[f(X_s(t))] - \min_{x \in \mathbb{R}^d} f(x) = \mathbb{E}[f(X_s(t)) - f(X_s(\infty))] + \underbrace{\mathbb{E}[f(X_s(\infty))] - \min_{x \in \mathbb{R}^d} f(x)}_{\text{final gap}},\] 
where the first part $\mathbb{E}[f(X_s(t)) - f(X_s(\infty))]$ converges linearly with the rate as 
\[
\lambda_s \asymp \exp\left(- \frac{2H_f}{s}\right)
\] and $H_f$ is the height difference (See the detail in~\cite[Section 6.2]{shi2020learning}). Also, the second part is the final gap, which can be bounded by the learning rate $s$.

Actually, the lr-dependent SDE~\eqref{eqn: lr-sde} is the overdamped limit of the hp-dependent SDE~\eqref{eqn: lr-momentum-sde}. Assume the learning rate $s \ll 1$ or $s \rightarrow 0$, then we can obtain the friction parameter 
\[
2\sqrt{\mu} = \frac{1 - \alpha}{1 + \alpha} \cdot \frac{2}{\sqrt{s}} \gg 1 \qquad \text{or} \qquad 2\sqrt{\mu} = \frac{1 - \alpha}{1 + \alpha} \cdot \frac{2}{\sqrt{s}}  \rightarrow \infty.
\]
Let us introduce two new variables $\tau$ and $\beta$ and substitute them as
\begin{equation}
\label{eqn: tau-t}
\tau = \frac{t}{2\sqrt{\mu}} = \frac{k\sqrt{s}}{2\sqrt{\mu}} = k \cdot \frac{ s(1 + \alpha)}{2(1 - \alpha)} = k \beta(s, \alpha). 
\end{equation}
With the basic calculations
\[
\frac{d^2 X}{dt^2} = \frac{1}{4\mu} \frac{d^2 X}{d\tau^2}, \qquad \frac{d X}{dt} = \frac{1}{2\sqrt{\mu}} \frac{dX}{d\tau}, \qquad \frac{dW(t)}{dt} = \frac{dW(2\sqrt{\mu} \tau)}{d (2\sqrt{\mu} \tau)} = \sqrt{\frac{1}{2\sqrt{\mu}}} \frac{dW(\tau)}{d\tau}, 
\]
we can obtain the equivalent form for the hp-dependent SDE~\eqref{eqn: lr-momentum-sde-standard} for the variable $\tau$ as
\[
\frac{1}{4\mu}\ddot{X} + \dot{X} + \nabla f(X)  = \sqrt{ \beta(s, \alpha) } \dot{W}.
\]
When the friction is overdamped, that is, the friction coefficient $\mu$ is oversized, then we have
\begin{equation}
\label{eqn: overdamped-lr-momentum-sde}
dX(\tau) = - \nabla f(X(\tau))d\tau + \sqrt{ \beta(s, \alpha) } dW(\tau),
\end{equation}
which is similar to the lr-dependent SDE~\eqref{eqn: lr-sde} and the only difference is that the coefficient of noise $\sqrt{s}$ is replaced by $\sqrt{\beta(s, \alpha)}$. Hence, recall~\cite{shi2020learning}, we find the equilibrium distribution for the lr-dependent SDE~\eqref{eqn: overdamped-lr-momentum-sde} is
\begin{equation}
\label{eqn: equi-alpha}
\mu_{s, \alpha} = \frac{1}{Z_{s, \alpha}} \exp\left( - \frac{2f}{\beta(s, \alpha)}  \right) = \frac{1}{Z_{s, \alpha}} \exp\left( - \frac{2f}{s} \cdot \frac{2(1-\alpha)}{1+\alpha} \right)
\end{equation}
and the rate of linear convergence is
\[
\lambda_{s, \alpha} \asymp \exp \left(- \frac{2H_f}{\beta(s, \alpha)} \right) = \exp \left(- \frac{2H_f}{s} \cdot \frac{2(1-\alpha)}{1+\alpha}\right),
\]
where $f$ is the potential, and $H_f$ is the height difference. Furthermore, we have
\begin{equation}
\label{eqn: conver-k}
\lambda_{s, \alpha} \tau  \asymp \exp \left(- \frac{2H_f}{\beta(s, \alpha)} \right)  \cdot k \beta(s, \alpha) =  \exp \left(- \frac{2H_f}{s} \cdot \frac{2(1-\alpha)}{1+\alpha}\right)  \cdot \frac{ 1 + \alpha}{2(1 - \alpha)} \cdot ks.
\end{equation}
Here, from~\eqref{eqn: equi-alpha} and~\eqref{eqn: conver-k}, we find the learning rate $s$ plays the same role as the lr-dependent SDE~\eqref{eqn: lr-sde}. Hence, we only need to discuss the effect of the momentum coefficient $\alpha$: 
\begin{itemize}
\item When $\alpha = 1/3$, then the equilibrium distribution $\mu_{s, \alpha}$ and the convergence rate $\lambda_{s, \alpha} \tau$ reduces to that in the lr-dependent SDE~\eqref{eqn: lr-sde}.
\item When $\alpha < 1/3$, then the equilibrium distribution $\mu_{s, \alpha}$ concentrates and the convergence rate $\lambda_{s, \alpha} \tau$ decreases sharply. Practically, we rarely adopt this setting. 
\item When $\alpha > 1/3$, then the equilibrium distribution $\mu_{s, \alpha}$ diverges, but the convergence rate $\lambda_{s, \alpha} \tau$  increases sharply. This is the setting we often adopt in practice. Here, we demonstrate that the parameters $\beta(s, \alpha)/s$ vary with the momentum coefficient $\alpha$, which are set as $0.5$, $0.9$ and $0.99$ in~\Cref{fig: alpha5999}. 
\begin{figure}[h!]
\centering
\begin{tabular}{|c|ccc|}
\hline 
$\alpha$                                           & 0.5 & 0.9 & 0.99 \\
\hline
$\frac{1+ \alpha}{2(1 - \alpha)}$     & $1.5$ & $9.5$ & $99.5$ \\
\hline
\end{tabular}
\caption{The parameters $\frac{1+ \alpha}{2(1 - \alpha)}$ varies with the momentum coefficient $\alpha$.}
\label{fig: alpha5999}
\end{figure}

\end{itemize}

Back to~\Cref{fig: msgd-deep}, the equilibrium distribution $\mu_{s, \alpha}$ is reflected in the height of the stable training error. When the equilibrium distribution $\mu_{s, \alpha}$ concentrates (diverges), the height of the stable training error decreases (increases). Generally speaking, when $\alpha = 0.5$, then $\frac{1+ \alpha}{2(1 - \alpha)} = 1.5$, there is no prominent change between the lr-dependent SDE and the hp-dependent SDE for the convergence rate $\lambda_{s, \alpha} \tau$ and the  height of the stable training error. However, the momentum coefficient is set as $\alpha = 0.9$, then $\frac{1+ \alpha}{2(1 - \alpha)} = 9.5$, the convergence rate $\lambda_{s, \alpha} \tau$ will increase sharply. Meanwhile, the stable training error will not increase so much.  Furthermore, if the convergence is still very slow with $\alpha = 0.9$, we can enlarge $\alpha > 0.9$ to accelerate the convergence rate $\lambda_{s, \alpha} \tau$. Still, the stable training error may rise so much that the final performance changes rarely compared with the initial value.

Based on this intuitive analysis, we know the coefficient of noise is $\sqrt{\beta(s, \alpha)}$ instead of $\sqrt{s}$ in the hp-dependent SDE~\eqref{eqn: lr-momentum-sde-standard} . Hence, the parameters $\frac{1 + \alpha}{2(1 - \alpha)}$ for the momentum coefficient $\alpha$ is set as the coefficient of learning rate $s$, which as a whole, conversely influences the linear rate of convergence and the equilibrium distribution. In Figure~\ref{fig: alpha5999}, we also find that the momentum coefficient $\alpha =0.9$ balances for the two reverse directions. This paper proposes rigorous proof corresponding to this intuition using modern mathematical techniques: hypocoercity and semi-classical analysis. 

\subsection{Related work}
\label{subsec: related-work}

Currently, the study of deep learning is a fashionable topic. Finding ways to tune the parameters is preoccupying the industry. The seminal work~\cite{bengio2012practical} discusses the significance of the hyperparameters in practice, which not only points out that the learning rate plays the single most crucial role but also suggests that the added momentum will lead to faster convergence in some cases. Moreover, in practice, as proposed in~\cite{he2016deep}, the classical Residual Networks adopt SGD with momentum,  not SGD, to obtain a sound performance for image recognition.

Recently, in the field of nonlinear optimization, there has been an emerging method called continuous-time approximation for discrete algorithms. By taking the approximating ODEs, we can simplify the discrete algorithms and use a modern form of analysis to obtain new characteristics, and the formation has not yet been discovered. This method starts to investigate the acceleration phenomenon generated by Nesterov's accelerated gradient methods~\cite{su2016differential, jordan2018dynamical}. Finally, it is solved in~\cite{shi2018understanding}   by introducing the high-resolution approximated differential equations based on the dimensional analysis from physics. 

In the stochastic setting, this approach has been recently pursued by various authors \cite{chaudhari2018deep,chaudhari2018stochastic,mandt2016variational,lee2016gradient,caluya2019gradient,li2017stochastic} to establish the various properties of stochastic optimization. As a notable advantage, the continuous-time perspective allows us to work without assumptions on the boundedness of the domain and gradients, as opposed to the older analyses of SGD (see, for example, \cite{hazan2008adaptive}).
Our work is partly motivated by the recent progress on Langevin dynamics, particularly for nonconvex settings~\cite{villani2009hypocoercivity, pavliotis2014stochastic, helffer2004quantitative,bovier2005metastability}. In Langevin dynamics, the learning rate $s$ in the hp-dependent SDE can be thought of as the temperature parameter and $2\sqrt{\mu}$ as a function of the learning rate $s$ and the momentum coefficient $\alpha$, can be thought of as the friction coefficient. 

\subsection{Organization}
\label{subsec: organization}

The remainder of the paper is structured as follows. In Section~\ref{sec: prelim}, we introduce the basic concepts and assumptions employed throughout this paper. Next, Section~\ref{sec: main-result} develops our main theorems and some comparisons analytically and numerically with SGD with momentum. 
Section~\ref{sec: qualitative} formally proves the linear convergence, and Section~\ref{sec: quantative} further quantifies the linear rate of convergence. Technical details of the proofs are deferred to the appendices. We conclude the paper in Section~\ref{sec: conclu} with a few directions for future research.

%% file: prelim.tex
\section{Preliminaries}
\label{sec: prelim}


%
%
%

Throughout this paper, we assume that the objective function $f$ is infinitely differentiable in $\mathbb{R}^d$; that is, $f \in C^{\infty}(\mathbb{R}^d)$. We use $\| \cdot \|$ to denote the standard Euclidean norm. Recall the confining condition for the objective $f$~(\cite[Definition 2.1]{shi2020learning}, also see~\cite{markowich1999trend, pavliotis2014stochastic}): $\lim_{\|x\| \rightarrow +\infty} f(x) = +\infty$ and $\exp(-2f/s)$ is integrable for all $s > 0$. This condition is quite mild and requires that the function grows sufficiently rapidly when $x$ is far from the origin. For convenience, we need to define some Hilbert spaces. For any $k = -1, 0, 1$, let $\langle \cdot, \cdot \rangle_{L^{2}(\mu_{s, \alpha}^k)}$ be  the inner product in the Hilbert space $L^{2}(\mu_{s, \alpha}^k)$, defined as
\[
\langle g_1, g_2 \rangle_{L^{2}(\mu_{s, \alpha}^k)} = \int_{\mathbb{R}^d} \int_{\mathbb{R}^d} g_1g_2 \mu_{s, \alpha}^{k} \dd v \dd x
\]
with any $g_1, g_2 \in L^{2}(\mu_{s, \alpha}^k)$. For any $g \in L^{2}(\mu_{s, \alpha}^k)$, the norm induced by the inner product is 
\[
\| g\|_{L^{2}(\mu_{s, \alpha}^k)} = \sqrt{\langle g, g \rangle_{L^{2}(\mu_{s, \alpha}^k)}}.
\]

Recall the hp-dependent SDE~\eqref{eqn: lr-momentum-sde-standard}. 
Similar as~\cite{shi2020learning}, the probability density $\rho_{s, \alpha}(t, \cdot, \cdot)$ of $( X_{s, \alpha}(t), V_{s, \alpha}(t) )$ evolves according to the~\textit{hp-dependent kinetic Fokker-Planck equation} 
\begin{equation}
\label{eqn: kinetic-FP}
\frac{\partial \rho_{s, \alpha}}{\partial t} = - v \cdot \nabla_x \rho_{s, \alpha} + \nabla_x f \cdot \nabla_v \rho_{s, \alpha}  + 2\sqrt{\mu} \nabla_v \cdot (v \rho_{s, \alpha} ) + \frac{\sqrt{s}}{2} \Delta_v \rho_{s, \alpha} ,
\end{equation}
with the initial condition $\rho_{s,\alpha}(0, \cdot, \cdot)$. Here, $\Delta_v \equiv \nabla_v \cdot \nabla_v$ is the Laplacian. For completeness, we derive the hp-dependent kinetic Fokker-Planck equation in~\Cref{subsec: kinetic-FP} from the hp-dependent SDE~\eqref{eqn: lr-momentum-sde-standard} by It\^o's formula and Chapman-Kolmogorov equation. If the objective $f$ satisfies the confining condition, then the hp-dependent kinetic Fokker-Planck equation~\eqref{eqn: kinetic-FP} admits a unique invariant Gibbs distribution that takes the form as
\begin{equation}
\label{eqn: gibbs-invariant-distr}
\mu_{s, \alpha} = \frac{1}{Z_{s, \alpha}} \exp\left( - \frac{2f + \|v\|^2}{\beta(s, \alpha)} \right),
\end{equation}
where the normalization factor is 
\[
Z_{s, \alpha} = \int_{\mathbb{R}^d}\int_{\mathbb{R}^d}  \exp\left( - \frac{2f + \|v\|^2}{\beta (s, \alpha) } \right) dv dx.
\] 
The proof of existence and uniqueness is shown in~\Cref{subsec: exist-unique-gibbs}. 

\paragraph{Villani conditions} To show the solution's existence and uniqueness to the hp-dependent kinetic Fokker-Planck equation~\eqref{eqn: kinetic-FP}, we still need to introduce the Villani conditions first. 
\begin{defn}[Villani Conditions~\cite{villani2009hypocoercivity}]
\label{defn: villani}
 A confining condition $f$ is said to satisfy the Villani conditions if 
 \begin{itemize}
 \item[\rm{(I)}]  When $\| x \| \rightarrow \infty$, we have 
                \[
                \frac{\|\nabla f\|^2}{s} - \Delta f \rightarrow \infty;
                \] 
 \item[\rm{(II)}] For any $x \in \mathbb{R}^d$, we have   
                \[
                \|\nabla^2 f\|_2 \leq C(1 + \|\nabla f\|),
                \]           
                where the matrix $2$-norm is $\| \nabla^2 f \|_2 = |\lambda_{\max}(\nabla^2 f) |$.
 \end{itemize}
\end{defn}

The Villani condition-(I) has been proposed in~\cite[Definition 2.3]{shi2020learning}, which says that the gradient has a sufficiently large squared norm compared with the Laplacian of the function.  Generally speaking, the polynomials of degrees no less than two satisfy the Villani condition-(I). For the hp-dependent kinetic Fokker-Planck equation~\eqref{eqn: kinetic-FP}, the existence and uniqueness of the solution are still the Villani condition-(II), which is called the relative bound in~\cite{villani2009hypocoercivity}. Actually, we will find the Fokker-Planck-Kramers operator $\mathscr{K}_{s, \alpha}$ has a compact resolvent under the two Villani conditions together~(See~\cite[Theorem 5.8, Remark 5.13(a)]{helffer2005hypoelliptic} and~\cite{li2012global}) in~\Cref{sec: quantative}.
Hence, taking any initial probability density $\rho_{s, \alpha}(0, \cdot, \cdot) \in L^{2}(\mu_{s, \alpha}^{-1})$, we have the following guarantee: 
\begin{lem}[Existence and uniqueness of the weak solution]
\label{lem: weak-soln-eu}
For any confining function $f$ satisfying the Villani Conditions (Definition~\ref{defn: villani}) and any initial $\rho_{s, \alpha}(0, \cdot, \cdot) \in L^{2}(\mu_{s, \alpha}^{-1})$, the hp-dependent SDE~\eqref{eqn: lr-momentum-sde-standard} admits a weak solution whose probability density in $C^{1}\left( [0, +\infty), L^{2}(\mu_{s, \alpha}^{-1}) \right)$ is the unique solution to the hp-dependent kinetic Fokker-Planck equation~\eqref{eqn: kinetic-FP}.
\end{lem}

\paragraph{Basics of Morse theory} Similar as~\cite{shi2020learning}, we also need to assume the objective function is a Morse function. Here, we will describe the basic concepts briefly. 
A point $x$ is called a \textit{critical point} if the gradient $\nabla f(x) = 0$. A function $f$ is said to be a \textit{Morse} function if for any critical point $x$, the Hessian $\nabla^{2} f(x)$ at $x$ is non-degenerate. Note also a~\textit{local minimum} $x^\bullet \in \mathcal{X}^\bullet$ is a critical point with all the eigenvalues of the Hessian at $x^\bullet$ positive and an \emph{index-1 saddle point} $x^\circ \in \mathcal{X}^\circ$ is a critical point where the Hessian at $x^\circ$ has exactly one negative eigenvalue, that is, $\eta_1(x^\circ) \geq \cdots \geq \eta_{d-1}(x^\circ) > 0,\; \eta_d(x^\circ) < 0$. 
Let $\mathcal{K}_{f(x^\circ)} := \big\{x \in \mathbb{R}^{d}: f(x) < f(x^\circ) \big\}$ denote the sublevel set at level $f(x^\circ)$. If the radius $r$ is sufficiently small, the set $\mathcal{K}_{f(x^\circ)} \cap \{x: \|x - x^\circ\| < r\}$ can be partitioned into two connected components, say $C_{1}(x^\circ, r)$ and $C_{2}(x^\circ, r)$. Therefore, we can introduce the most important concept --- \textit{index-$1$ separating saddle} as follows.

\begin{defn}[\textbf{Index-$1$ Separating Saddle}]\label{defn: separating-saddle}
Let $x^\circ$ be an index-$1$ saddle point and $r > 0$ be sufficiently small. If $C_1(x^\circ, r)$ and $C_{2}(x^\circ, r)$ are contained in two different (maximal) connected components of the sublevel set $\mathcal{K}_{f(x^\circ)}$, we call $x^\circ$ an index-$1$ \textit{separating} saddle point.

\end{defn}
%
%
\noindent Intuitively speaking, the index-1 separating saddle point $x^\circ$ is the bottleneck of any path connecting the two local minima. More precisely, along a path connecting $x_1^\bullet$ and $x_2^\bullet$, by definition the function $f$ must attain a value that is at least as large as $f(x^\circ)$. For the detail about the basics of Morse theory, please readers refer to~\cite[Section 6.2]{shi2020learning}.

%% file: result.tex
\section{Main Results}
\label{sec: main-result}

In this section, we state our main results. Briefly, for SGD with momentum in its continuous formulation,  the hp-dependent SDE,  we show that the expected excess risk converges linearly to stationarity and estimate the final excess risk by the hyperparameters in~\Cref{sec:linear-convergence}. Furthermore, we derive a quantitative expression of the rate of linear convergence in~\Cref{sec:rate-line-conv}. Finally, we carry the continuous-time convergence guarantees to the discrete case in~\Cref{subsec: discretization}.

\subsection{Linear convergence}
\label{sec:linear-convergence}

In this subsection, we are concerned with the expected excess risk, $\mathbb{E}[f(X_{s, \alpha}(t))] - f^\star$. Recall that $f^\star = \inf_{x \in \mathbb{R}^d} f(x)$. 

\begin{mainthm}\label{thm: continuous-qualitative}
Let $f$ be confined and satisfy the Villani conditions. Then, there exists $\lambda_{s, \alpha} > 0$ for any learning rate $s > 0$ and any momentum coefficient $\alpha \in (0, 1)$ such that the expected excess risk satisfies
\begin{equation}\label{eqn: continuous-qualitative}
\E [f(X_{s, \alpha}(t))] - f^\star  \le \epsilon(s, \alpha) + D(s, \alpha) e^{-\lambda_{s, \alpha}  t},
\end{equation}
for all $t \ge 0$. Here $\epsilon(s, \alpha) = \epsilon(s, \alpha; f) \ge 0$ increases strictly for the mixing parameter $\beta(s, \alpha)$ and depends only on the objective function $f$, and $D(s, \alpha) = D(s, \alpha; f, \rho) \ge 0$ depends only on $s, \alpha, f$, and the initial distribution $\rho_{s, \alpha}(0, \cdot, \cdot)$.

\end{mainthm}

Similar to as~\cite{shi2020learning}, the proof of this theorem is based on the following decomposition of the expected excess risk:
\[
\mathbb{E}[f(X_{s, \alpha}(t))] - f^\star = \mathbb{E}[f(X_{s, \alpha}(t))] - \mathbb{E}[f(X_{s, \alpha}(\infty))] + \mathbb{E}[f(X_{s, \alpha}(\infty))]  - f^\star,
\]
where $\mathbb{E}[f(X_{s, \alpha}(\infty))]$ denotes $\mathbb{E}_{x \sim \mu_{s, \alpha}}[f(x)]$ in light of the fact that $X_{s, \alpha}(t)$ converges weakly to $\mu_{s, \alpha}$ as $t \rightarrow +\infty$ (see Lemma~\ref{lem: converge}). The question is thus separated into quantifying how fast $\mathbb{E}[f(X_{s, \alpha}(t))]$ converges weakly to $\mathbb{E}[f(X_{s, \alpha}(\infty))]$ as $t \rightarrow +\infty$ and how the expected excess risk at stationarity $\mathbb{E}[f(X_{s, \alpha}(\infty))] - f^\star$ depends on the hyperparameters. The following two propositions address these two questions. Recall that $\rho_{s, \alpha}(0, \cdot, \cdot) \in L^2(\mu_{s, \alpha}^{-1})$ is the probability density of the initial iterate in SGD with momentum. 

\begin{prop}
\label{prop: linear-convergence}
Under the assumptions of Theorem~\ref{thm: continuous-qualitative}, there exists $\lambda_{s, \alpha} > 0$ for any learning rate $s$ and any momentum coefficient $\alpha \in (0, 1)$ such that
\[
\left| \mathbb{E}[f(X_{s, \alpha}(t))] - \mathbb{E}[f(X_{s, \alpha}(\infty))] \right| \leq C(s, \alpha) \left\| \rho_{s, \alpha}(0, \cdot, \cdot) - \mu_{s, \alpha} \right\|_{ L^2(\mu_{s, \alpha}^{-1}) } e^{- \lambda_{s, \alpha} t},
\]
for any $t \geq 0$, where the constant $C(s, \alpha)$ depends only $s$, $\alpha$ and $f$, and where 
\[
\left\| \rho_{s, \alpha}(0, \cdot, \cdot) - \mu_{s, \alpha} \right\|_{ L^2(\mu_{s, \alpha}^{-1}) }  = \left( \int_{\mathbb{R}^d} \int_{\mathbb{R}^d} \left( \rho_{s, \alpha}(0, \cdot, \cdot) - \mu_{s, \alpha} \right)^{2} \mu_{s, \alpha}^{-1} dvdx \right)^{\frac{1}{2}}
\]
measures the gap between the initialization and the Gibbs invariant distribution.
\end{prop}

Loosely speaking, it takes $O(1/\lambda_{s, \alpha})$ to converge to stationarity from the beginning. In Theorem~\ref{thm: continuous-qualitative}, $D(s, \alpha)$ can be set to $C(s, \alpha) \| \rho_{s, \alpha}(0, \cdot, \cdot)  - \mu_{s, \alpha} \|_{L^2( \mu_{s, \alpha}^{-1}) }$. Notably, the proof of Proposition~\ref{prop: linear-convergence} shall reveal that $C(s, \alpha)$ increases as the learning rate $s$ increases. Turning to the analysis of the second term, $\mathbb{E}[f(X_{s, \alpha}(\infty))] - f^\star$, we write henceforth $\epsilon(s, \alpha) := \mathbb{E}[f(X_{s, \alpha}(\infty))] - f^\star$.

\begin{prop}
\label{prop: final-gap}
Under the assumptions of Theorem~\ref{thm: continuous-qualitative}, the expected excess risk at stationarity, $\epsilon(s, \alpha)$, is a strictly increasing function $\beta(s, \alpha)$. Moreover, for any $S>0$, there exists a constant $A$ that depends only on $S$ and $f$ and satisfies 
\[
\epsilon(s, \alpha) \equiv \mathbb{E}\left[ X_{s, \alpha}(\infty) \right] - f^\star \leq A\beta(s, \alpha) = \frac{A(1 + \alpha)}{2(1 - \alpha)} \cdot s, 
\]
for any learning rate $0 < s \leq S$.
\end{prop}

The two propositions are proved in \Cref{sec: qualitative}. The proof of Theorem~\ref{thm: continuous-qualitative} is a direct consequence of Proposition~\ref{prop: linear-convergence} and Proposition~\ref{prop: final-gap}. More precisely, the two propositions taken together give
\begin{equation}\label{eq:as_strong}
\mathbb{E}  f(X_{s, \alpha}(t)) - f^\star  \le O\left( \frac{1 + \alpha}{2(1 - \alpha)} \cdot s\right) + C(s, \alpha) e^{-\lambda_{s, \alpha}  t},
\end{equation}
for a bounded learning rate $s$.
%
%
%
%
%
%
%
%
The following result gives the iteration complexity of SGD with momentum in its continuous-time formulation.
\begin{coro}\label{coro:iter_epsi}
Under the assumptions of Theorem~\ref{thm: continuous-qualitative}, for any $\epsilon > 0$, if the learning rate $s \le \min\{ \epsilon/A \cdot (1-\alpha) / (1 + \alpha), S\}$ and $t \ge \frac1{\lambda_{s, \alpha}} \log \frac{2C(s, \alpha) \left\| \rho_{s, \alpha}(0, \cdot, \cdot) - \mu_{s, \alpha} \right\|_{  L^2(\mu_{s, \alpha}^{-1})  } }{\epsilon}$, then
\[
\mathbb{E} \left[ f(X_{s, \alpha}(t)) \right] - f^\star \le \epsilon.
\]
\end{coro}

\subsection{The rate of linear convergence}
\label{sec:rate-line-conv}

Now, we turn to the key issue of understanding how the linear rate $\lambda_{s, \alpha}$ depends on the hyperparameters, the learning rate $s$, and the momentum coefficient $\alpha$. Here, we propose an explicit expression for the linear rate $\lambda_{s, \alpha}$ 
to interpret this.

\begin{mainthm}\label{thm: continuous-quantative}
In addition to the assumptions of Theorem~\ref{thm: continuous-qualitative}, assume that the objective $f$ is a Morse function and has at least two local minima.
Then the constant $\lambda_{s, \alpha}$ in~\eqref{eqn: continuous-qualitative} satisfies
\begin{equation}\label{eqn: main-lambda-nonconvex}
\lambda_{s,\alpha} = \frac{v(\gamma + o(s))}{\sqrt{\mu} + \sqrt{\mu + v}} e^{- \frac{2H_{f}}{\beta(s, \alpha)}} = \frac{v(\gamma + o(s))}{\sqrt{\mu} + \sqrt{\mu + v}} e^{- \frac{2H_{f}}{s} \cdot \frac{2(1 - \alpha)}{1 + \alpha}} ,
\end{equation}
for $0 < s \le s_0$, where $s_0 >0, \alpha > 0$, $H_{f} > 0$ are constants completely depending \textit{only} on $f$ and $-v$ is the unique negative eigenvalue of the Hessian at the highest index-$1$ separating saddle. 

\end{mainthm}

Recall~\cite[Theorem 2]{shi2020learning}, for the lr-dependent SDE~\eqref{eqn: lr-sde}, the exponential decay constant is obtained as
\begin{equation}
\label{eqn: lambda-s1}
\lambda_{s} = v(\gamma + o(s)) e^{-\frac{2H_f}{s}}.
\end{equation}
Here, we first come to analyze the ratio of two exponential decays, $\lambda_{s, \alpha}$ in~\eqref{eqn: main-lambda-nonconvex} and $\lambda_s$ in~\eqref{eqn: lambda-s1}, as 
\[
\frac{\lambda_{s, \alpha}}{\lambda_{s}} = \frac{1}{\sqrt{\mu} + \sqrt{\mu + v}}e^{- \frac{2H_{f}}{s} \cdot \frac{1 - 3\alpha }{1 + \alpha}}.
\]
 Then, we compute the ratio of the exponential decay constants in~\eqref{eqn: main-lambda-nonconvex} with different learning rates, $s_1$ and $s_2$, as
\[
\frac{\lambda_{s_1, \alpha}}{\lambda_{s_2, \alpha}} = \sqrt{\frac{s_1}{s_2}} \left( \frac{\lambda_{s_1}}{\lambda_{s_2}} \right)^{\frac{2(1 - \alpha)}{1+\alpha}}.
\] 
For any $\alpha > 1/3$, when $s, s_1, s_2 \rightarrow 0$, we can obtain
\begin{equation}
\label{eqn: ratio-1}
\frac{\lambda_{s, \alpha}}{\lambda_{s}}  \asymp \frac{1+ \alpha}{1 - \alpha} \cdot 2\sqrt{s}e^{\frac{2H_{f}}{s} \cdot \frac{3\alpha - 1}{1 + \alpha}} \rightarrow +\infty,
\end{equation}
and there exist two constants $C_1$ and  $C_2$ in $(0, 1)$ such that
\begin{equation}
\label{eqn: ratio-2}
\left( \frac{\lambda_{s_1}}{\lambda_{s_2}} \right)^{C_1} \leq \frac{\lambda_{s_1, \alpha}}{\lambda_{s_2, \alpha}} \leq \left( \frac{\lambda_{s_1}}{\lambda_{s_2}} \right)^{C_2}.
\end{equation}
From~\eqref{eqn: ratio-1}, we can find when the learning rate $s$ is sufficiently small, $\lambda_{s, \alpha}$ is larger than $\lambda_s$. In other words, the SGD with momentum converges faster than SGD. Furthermore, from~\eqref{eqn: ratio-2}, the ratio $\lambda_{s_1, \alpha} / \lambda_{s_2, \alpha}$ is weaker than $\lambda_{s_1} / \lambda_{s_2}$, that is, $\lambda_{s_1, \alpha} / \lambda_{s_2, \alpha}$ is closer to $1$ than  $\lambda_{s_1} / \lambda_{s_2}$. In~\cite{shi2020learning}, we show the exponential decay constant $\lambda_{s}$ changes sharply with the learning rate $s$. Here, we can find the exponential decay constant $\lambda_{s, \alpha}$ in SGD with momentum is more robust with the learning rate $s$ than $\lambda_s$.
%

\begin{figure}[htb!]
\centering
\begin{minipage}[t]{0.45\linewidth}
\centering
\includegraphics[scale=0.14]{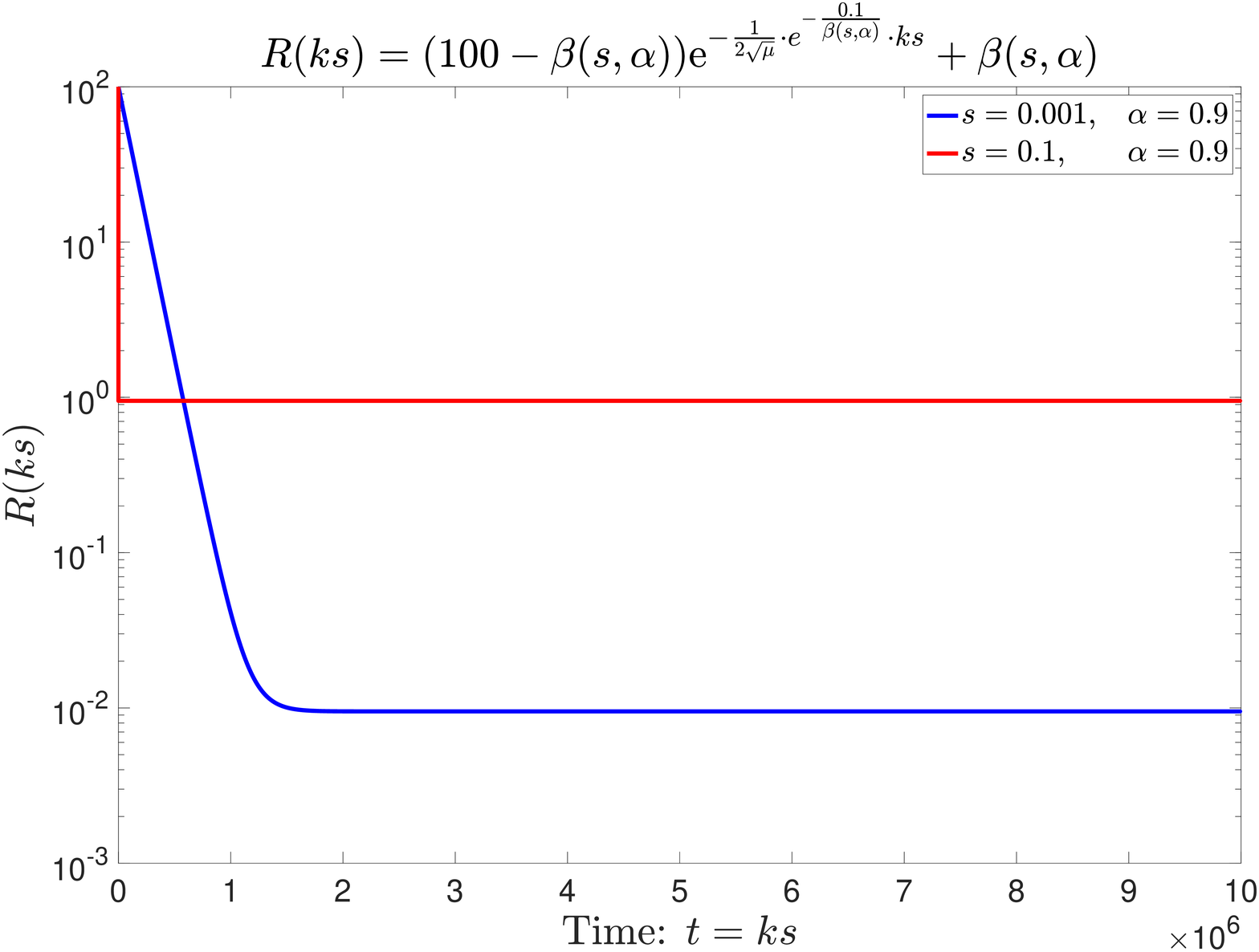}
\end{minipage}
\begin{minipage}[t]{0.45\linewidth}
\centering
\includegraphics[scale=0.14]{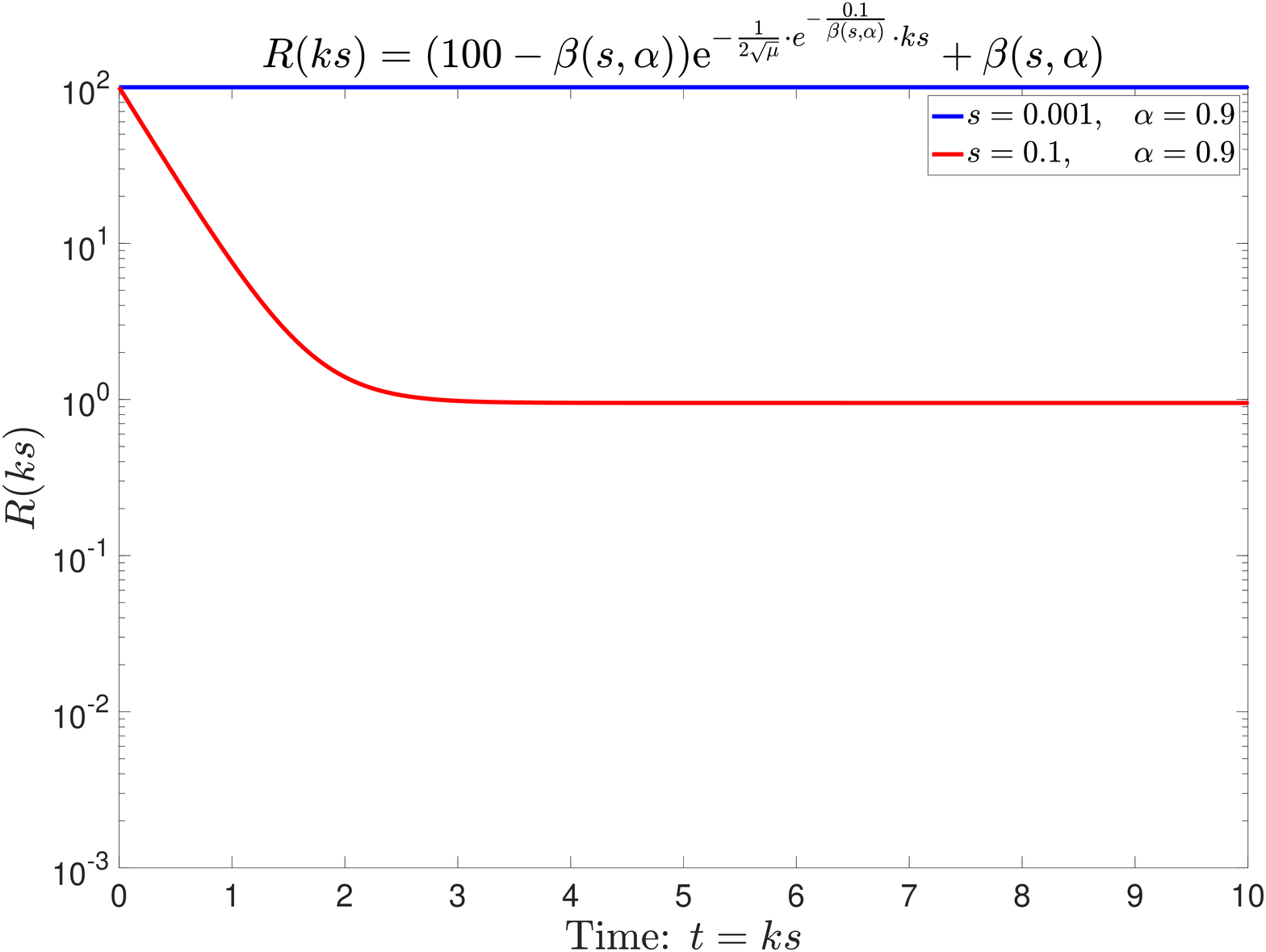}
\end{minipage}
\caption{\small Idealized risk function of the form $R(t) = \left( 100 - \beta(s, \alpha) \right) e^{- \frac{1}{2\sqrt{\mu}} \cdot e^{-\frac{0.1}{\beta(s, \alpha)}} t} + \beta(s, \alpha)$ with the identification $t = ks$, which is adapted from \eqref{eqn: continuous-qualitative}. Similar as~\cite[Figure 7]{shi2020learning}, the learning rate is selected as $s = 0.1$ and $0.001$. The right plot is a locally enlarged image of the left.
} 
\label{fig: exp2_s}
\end{figure}

\paragraph{Numerical Demonstration}
We demonstrate a numerical comparison based on the analysis and description above. Recall~\cite[Figure 7]{shi2020learning}, we compare the iteration number of lr-dependent SDE for the learning rates $s = 0.001$ and $s = 0.1$ as
\[
k_{0.001} \approx 2.5 \times 10^{47}, \quad  k_{0.1} \approx 2.5 \times 10^2, \quad \text{and} \quad k_{0.001}/k_{0.1} \approx 10^{45}.
\]
\noindent Here, we consider the idealized risk function for hp-dependent SDE~\eqref{eqn: lr-momentum-sde-standard} with the form \[
R(t) = \left( 100 - \beta(s, \alpha) \right) e^{- \frac{1}{2\sqrt{\mu}} \cdot e^{-\frac{0.1}{\beta(s, \alpha)}} t} + \beta(s, \alpha),\] shown in Figure~\ref{fig: exp2_s}. With the basic calculation, we can obtain that
\[
k_{0.001,0.9} \approx 1.5 \times 10^{9}, \quad  k_{0.1,0.9} \approx 2.5 \times 10^1, \quad \text{and} \quad k_{0.001, 0.9}/k_{0.1, 0.9}  \approx 6 \times 10^{7}.
\]
Apparently, when the momentum coefficient is set as $\alpha = 0.9$, $k_{0.001,0.9} < k_{0.001}$ and $k_{0.1,0.9} < k_{0.1}$, that verifies the iteration number for SGD with momentum is far less than that for SGD. Moreover, the ratio for the iterative number $k_{0.001, 0.9}/k_{0.1, 0.9}$ is also far less than $k_{0.001}/k_{0.1}$, that is, the iteration number in SGD with momentum is more robust than SGD. 


\subsection{Discretization}
\label{subsec: discretization}

This subsection presents the results developed from the continuous perspective of the discrete regime. For the discrete algorithms, we still need to assume $f$ to be $L$-smooth, that is, the gradient of $f$ is $L$-Lipschitz continuous in the sense that $\| \nabla f(x) - \nabla f(y) \| \leq L \| x - y \|$ for all $x, y \in \mathbb{R}^d$. Therefore, we can restrict the learning rate $s$ to be no larger than $1/L$. The following proposition is the key tool for translation to the discrete regime.
\begin{prop}
\label{prop: approx}
For any $L$-smooth objective $f$ and any initialization $X_{s, \alpha}(0)$ drawn from a probability density $\rho_{s, \alpha}(0, \cdot, \cdot) \in L^{2}(\mu_{s, \alpha}^{-1})$, the~hp-dependent SDE~\eqref{eqn: lr-momentum-sde-standard} has a unique global solution $X_{s, \alpha}$  in expectation; that is, $\E[X_{s, \alpha}(t)]$ as a function of $t$ in $C^{1}([0, + \infty); \mathbb{R}^{d})$ is unique. Moreover, there exists  $B(T) > 0$ such that the SGD with momentum iterates $x_{k}$ satisfy
\[
\max_{0 \leq k \leq T/s} \left| \E f(x_k) - \E f(X_s(ks)) \right| \leq B(T) s,
\]
for any fixed $T > 0$.
\end{prop}

We note a sharp bound on $B(T)$ in~\cite{bally1996law}. For completeness, we also remark that the convergence can be strengthened to the strong sense:
\[
 \max_{0 \leq k \leq T/s} \E   \left\| x_k - X_s(ks)  \right\| \leq B'(T)s.
\]
This result has appeared in \cite{mil1975approximate, talay1982analyse, pardoux1985approximation, talay1984efficient, kloeden1992approximation} and we provide a self-contained proof in Appendix~\ref{subsec: proof-discrete}. We now state the main result of this subsection.
\begin{mainthm}\label{thm: main1}
In addition to the assumptions of Theorem~\ref{thm: continuous-qualitative}, assume that $f$ is $L$-smooth. Then, the following two conclusions hold:
\begin{enumerate}
\item[(a)]
For any $T > 0$ and any learning rate $0 < s \le 1/L$, the iterates of SGD with momentum   satisfy
\begin{equation}\label{eqn: final-estimate-sgd}
\E f(x_k) - f^{\star} \leq \left( \frac{A(1 + \alpha)}{2(1 - \alpha)} + B(T) \right)s + C \left\| \rho - \mu_{s, \alpha} \right\|_{L^2(\mu_{s, \alpha}^{-1})} \mathrm{e}^{- s\lambda_{s, \alpha} k},
\end{equation}
for all $k \le T/s$, where $\lambda_{s, \alpha}$ is the exponential decay constant in~\eqref{eqn: continuous-qualitative}, $A$ as in Proposition~\ref{prop: final-gap} depends only on $1/L$ and $f$, $C = C_{1/L}$ is as in Proposition~\ref{prop: linear-convergence}, and $B(T)$ depends only on the time horizon $T$ and the Lipschitz constant $L$. 

\item[(b)] If $f$ is a Morse function with at least two local minima, with $\lambda_{s, \alpha}$ appearing in \eqref{eqn: final-estimate-sgd} being given by \eqref{eqn: main-lambda-nonconvex}.
\end{enumerate}

\end{mainthm}

Theorem~\ref{thm: main1} follows as a direct consequence of Theorem~\ref{thm: continuous-qualitative} and Proposition~\ref{prop: approx}. Note that the second part of Theorem~\ref{thm: main1} is simply a restatement of Theorem~\ref{thm: continuous-quantative}. 
Moreover, we also mention that the dimension parameter $d$ is not essential for characterizing the linear rate of convergence.

%% file: qualitative.tex
\section{Proof of the Linear Convergence}
\label{sec: qualitative}

In this section, we prove Proposition~\ref{sec:linear-convergence} and Proposition~\ref{prop: final-gap}, which lead to a complete proof of Theorem~\ref{thm: continuous-qualitative}.

\subsection{Linear operators and convergence}                                         
\label{subsec: linear-operator-convergence}                                           
Before proving Proposition~\ref{sec:linear-convergence}, we first take a simple analysis for the linear operators derived from the hp-dependent kinetic Fokker-Planck equation~\eqref{eqn: kinetic-FP}.  Then, we point out that only the Villani condition-$\text{(I)}$ and the Poincar\'e inequality cannot work here. To obtain the estimate for the hp-dependent kinetic Fokker-Planck equation~\eqref{eqn: kinetic-FP}, we still need to introduce the Villani condition-$\text{(II)}$ and demonstrate a vital inequality, which is named the relative bound~\cite{villani2009hypocoercivity}.

 \paragraph{Basic Property of Linear Operators}

Similar to the transformation used in~\cite[Section 5]{shi2020learning},  we obtain the time-evolving probability density over the Gibbs invariant measure given as
\[
h_{s, \alpha}(t, \cdot, \cdot) = \rho_{s, \alpha}(t, \cdot, \cdot) \mu_{s, \alpha}^{-1} \in C^1\left( [0, +\infty), L^2(\mu_{s, \alpha}) \right),
\]
which allows us to work in the space $L^{2}(\mu_{s, \alpha})$ in place of $L^{2}(\mu_{s, \alpha}^{-1})$. 
It is not hard to show that $h_{s, \alpha}$ satisfies the following differential equation
\begin{equation}
\label{eqn: kinetic-equal1}
\frac{\partial h_{s, \alpha}}{\partial t} = - \mathscr{L}_{s, \alpha} h_{s, \alpha}, 
\end{equation}
with the initial  $h_{s, \alpha}(0, \cdot, \cdot) = \rho_{s, \alpha}(0, \cdot, \cdot) \mu_{s, \alpha}^{-1} \in L^{2}(\mu_{s, \alpha})$, where the linear operator is defined as
\begin{equation}
\label{eqn: kramers-fp}
\mathscr{L}_{s, \alpha} = v \cdot \nabla_x  - \nabla_x f \cdot \nabla_v + 2\sqrt{\mu} v \cdot \nabla_v  - \frac{\sqrt{s}}{2} \Delta_v. 
\end{equation}
Simple observation tells us that the linear operator~\eqref{eqn: kramers-fp} can be separated into two parts 
\begin{equation}
\label{eqn: kramers-fp-separate}
\mathscr{L}_{s, \alpha} = \mathscr{T} + \mathscr{D}_{s, \alpha} 
\end{equation}
where the first part $ \mathscr{T}  =   v \cdot \nabla_x  - \nabla_x f \cdot \nabla_v $ is named as~\textit{transport operator} and the second part $\mathscr{S}_{s, \alpha}  =  2\sqrt{\mu} v \cdot \nabla_v  - \frac{\sqrt{s}}{2} \Delta_v $ is named as~\textit{diffusion operator}. 

Let $[\cdot, \cdot]$ be the commutator operation and the two new operators 
\begin{equation}
\label{eqn: gradient-v-x}
\mathscr{A}_{s, \alpha} = \sqrt[4]{\frac{s}{4}}\nabla_v \quad \text{and} \quad \mathscr{C}_{s, \alpha} = \sqrt[4]{\frac{s}{4}}\nabla_x,
\end{equation}
then we can obtain the basic facts described in the following lemma. 

\begin{lem}
\label{lem: basic-operator}
\; In the Hilbert space $L^{2}(\mu_{s, \alpha})$, with the notation of the linear operators above, we have 
\begin{enumerate}[label=\text{(\roman*)}]
\item The conjugate operators of the linear operators, $\mathscr{A}_{s, \alpha}$ and $\mathscr{C}_{s, \alpha}$, are
         \begin{equation}
         \label{eqn: conjugate-A}
         \mathscr{A}_{s, \alpha}^\star =   \sqrt[4]{\frac{s}{4}} \left( - \nabla_v + \frac{2v}{\beta} \right) \quad \text{and} \quad \mathscr{C}_{s, \alpha}^\star =   \sqrt[4]{\frac{s}{4}} \left( - \nabla_x + \frac{2\nabla_x f}{\beta} \right).
         \end{equation}
         Moreover, the diffusion operator is $\mathscr{D}_{s, \alpha} =  \mathscr{A}_{s, \alpha}^\star \mathscr{A}_{s, \alpha}$. 

\item  The linear operators $\mathscr{A}_{s, \alpha}$ and $\mathscr{A}^\star_{s, \alpha}$ commutes with $\mathscr{C}_{s, \alpha}$. Also, the commutator between  $\mathscr{A}_{s, \alpha}$ and $\mathscr{A}^\star_{s, \alpha}$ is 
          \begin{equation}
          \label{eqn: commutator-A-A-star}
          [\mathscr{A}_{s, \alpha}, \mathscr{A}_{s, \alpha}^\star] =  2\sqrt{\mu} \mathbf{I}_{d \times d}.
          \end{equation} 

\item The transport operator $\mathscr{T}$ is anti-symmetric. The commutator between $\mathscr{A}_{s, \alpha}$ and $\mathscr{T}$ is
          \begin{equation}
          \label{eqn: commutator-A-T}
          [\mathscr{A}_{s, \alpha}, \mathscr{T}] = \mathscr{C}_{s, \alpha}
          \end{equation}                  
          and that between $\mathscr{C}_{s, \alpha}$ and $\mathscr{T}$ is
          \begin{equation}
          \label{eqn: commutator-A-T}
          [\mathscr{C}_{s, \alpha}, \mathscr{T}] = - \nabla_x^2 f \cdot \mathscr{A}_{s, \alpha}.
          \end{equation}

%
%
%

\end{enumerate}
\end{lem}

The proof is only based on the basic operations and integration by parts, which is shown in Appendix~\ref{subsec: proof-lemma41}. 
%
%
%
%
%
%

\paragraph{Convergence to Gibbs invariant measure} 
Similarly as~\cite{shi2020learning}, we need to claim the function in $L^2(\mu_{s, \alpha}^{-1})$ is integrable, that is, $L^2(\mu_{s, \alpha}^{-1})$ is a subset of $L^1(\mathbb{R}^d)$. 

\begin{lem}\;
\label{lem: l2-l1}
Let $f$ satisfy the confining condition. Then, $L^2(\mu_{s, \alpha}^{-1}) \subset L^1(\mathbb{R}^d)$. 
\end{lem}



\noindent The proof of Lemma~\ref{lem: l2-l1} is simple and shown in Appendix~\ref{subsec: proof-l2-l1}. With Lemma~\ref{lem: basic-operator}, we claim the basic fact as the following lemma. 

\begin{lem}\;
\label{lem: equivlent-h}
The linear operator $\mathscr{L}_{s, \alpha}$ is nonpositive in $L^{2}(\mu_{s, \alpha})$. Explicitly, for any $g \in L^{2}(\mu_{s, \alpha})$, the linear operator  $\mathscr{L}_{s, \alpha}$ obeys 
\[
\left\langle \mathscr{L}_{s, \alpha}g, g \right\rangle_{L^2(\mu_{s, \alpha})} = \left\langle \mathscr{D}_{s, \alpha}g, g \right\rangle_{L^2(\mu_{s, \alpha})} = \| \mathscr{A}_{s, \alpha} g\|_{L^2(\mu_{s, \alpha})}^2 =  \frac{\sqrt{s}}{2} \int_{\mathbb{R}^d} \int_{\mathbb{R}^d}  \| \nabla_v g \|^2 \mu_{s, \alpha} \dd v \dd x.
\]

%
\end{lem}

With Lemma~\ref{lem: equivlent-h}, we can show the solution to the hp-dependent SDE~\eqref{eqn: kinetic-FP} converges to the Gibbs invariant measure in terms of the dynamics of its probability densities over time.


\begin{lem}\label{lem: converge}\;
Let $f$ satisfy the confining condition and denote the initial distribution as $\rho_{s, \alpha}(0, \cdot, \cdot) \in L^{2}(\mu_{s, \alpha}^{-1})$. Then, the unique solution $\rho_{s, \alpha}(t, \cdot, \cdot) \in C^{1} \left( [0, +\infty), L^{2}(\mu_{s, \alpha}^{-1}) \right)$ to the hp-dependent Fokker-Planck equation~\eqref{eqn: kinetic-FP} converges in $L^{2}(\mu_{s,\alpha}^{-1})$ to the Gibbs invariant measure $\mu_{s,\alpha}$, which is specified by~\eqref{eqn: gibbs-invariant-distr}.
\end{lem}

\begin{proof}[Proof of Lemma~\ref{lem: converge}] 
Taking a derivative for the $L^2$ distance between the time involving probability density $ \rho_{s, \alpha}(t, \cdot, \cdot)$ and the Gibbs invariant measure $ \mu_{s, \alpha}$, we have
\begin{align*}
\frac{d}{d t} \| \rho_{s, \alpha}(t, \cdot, \cdot) - \mu_{s, \alpha} \|_{L^2(\mu_{s, \alpha}^{-1})}^{2} & = \frac{d}{d t} \| h_{s, \alpha}(t, \cdot, \cdot) - 1 \|_{L^2(\mu_{s, \alpha})}^{2} \\
                                                                                                                                   & = \frac{d}{d t} \int_{\mathbb{R}^d} \int_{\mathbb{R}^d} \left(h_{s, \alpha}(t, x, v) - 1\right)^2 \mu_{s, \alpha} \dd v\dd x  \\
                                                                                                                                   & = \int_{\mathbb{R}^d} \int_{\mathbb{R}^d} -\mathscr{L}_{s, \alpha} \left(h_{s, \alpha}(t, x, v) - 1\right) \left( h_{s, \alpha}(t, x, v)  - 1 \right) \mu_{s, \alpha} \dd v \dd x, 
\end{align*}
The last equality is due to the equation~\eqref{eqn: kinetic-equal1}. Next, by use of Lemma~\ref{lem: equivlent-h}, we can proceed to
\begin{align}
\label{eqn: only-v-inequality}
\frac{\dd}{\dd t} \| \rho_{s, \alpha}(t, \cdot, \cdot) - \mu_{s, \alpha} \|_{L^2(\mu_{s, \alpha}^{-1})}^{2}  = - \frac{\sqrt{s}}{2} \int_{\mathbb{R}^d} \int_{\mathbb{R}^d} \| \nabla_v h_{s, \alpha}(t, x, v) \|^2 \mu_{s, \alpha}  \dd v\dd x \leq 0.
\end{align}
Thus, $ \| \rho_{s, \alpha}(t, \cdot, \cdot) - \mu_{s, \alpha} \|_{L^2(\mu_{s, \alpha}^{-1})}^{2}$ is decreasing strictly and  asymptotically towards the equilibrium state 
\[
\int_{\mathbb{R}^d} \int_{\mathbb{R}^d} \| \nabla_v h_{s, \alpha}(t, x,v) \|^2 d vd x = 0.
\]
This equality holds, however, only if $h_{s, \alpha}(t, \cdot, \cdot)$ is constant about $v$. Back to the differential equation~\eqref{eqn: kinetic-equal1}, we have
\[
\frac{\partial h_{s, \alpha}}{\partial t} = - v \cdot \nabla_x h_{s, \alpha}, 
\]
of which the solution is  $h_{s,\alpha} = h_{s,\alpha}(0, x + tv)$. Furthermore, we can obtain $\nabla_v h_{s, \alpha} = t \nabla_x h_{s, \alpha} $. With the time $t$'s arbitrarity, we know $\nabla_x h_{s, \alpha} = 0$. Therefore, we obtain that $h_{s, \alpha}(t, \cdot, \cdot)$ is constant.  The fact that both $\rho_{s,\alpha}(t,\cdot,\cdot)$ and $\mu_{s,\alpha}$ are probability densities implies that $h_{s,\alpha}(t, \cdot) \equiv 1$; that is, $\rho_{s,\alpha}(t,\cdot,\cdot) \equiv \mu_{s, \alpha}$.
\end{proof} 

\paragraph{Failure of Poincar\'e inequality} Recall the Villani Condition-$\text{(I)}$ in Definition~\ref{defn: villani}
\[
\frac{\| \nabla f \|^2}{s} - \Delta f \rightarrow +\infty,  \qquad \text{with} \quad \|x\| \rightarrow +\infty.
\]
In~\cite[Lemma 5.4]{shi2020learning}, we obtain the Poincar\'e inequality to derive the linear convergence, which is based on the technique in~\cite[Theorem A.1]{villani2009hypocoercivity}. Here, for the differential equation~\eqref{eqn: kinetic-equal1}, what we consider is not only the potential $f(x)$ itself but is the Hamiltonian.
\[
H(x, v) = f(x) + \frac{1}{2} \|v\|^2.  
\]
With the new norm $\| \nabla H \|^2 = \| \nabla_x H\|^2 + \| \nabla_v H \|^2$, the Villani condition-$\text{(I)}$ directly leads to
\[
\frac{ \| \nabla H \|^2 }{\beta(s, \alpha)} - \Delta H = \frac{ \| \nabla_x f \|^2 + \|v\|^2 }{\beta(s, \alpha)}  - \Delta_x f - d \rightarrow + \infty \qquad \text{with}\quad \|x\|^2 + \|v\|^2 \rightarrow +\infty.
\]
Based on the same technique in~\cite[Theorem A.1]{villani2009hypocoercivity}, we can obtain the following Poincar\'e inequality
\begin{equation}
\label{eqn: poincare}
\left\| h_{s, \alpha} - 1 \right\|^2_{L^2(\mu_{s, \alpha})}
 \leq  
\int_{\mathbb{R}^d} \int_{ \mathbb{R}^d}\left( \| \nabla_x h_{s, \alpha} \|^2 +  \| \nabla_v h_{s, \alpha} \|^2  \right) \mu_{s, \alpha} d x d v.
\end{equation}
Different from~\cite{shi2020learning}, this inequality above cannot connect with the equation~\eqref{eqn: only-v-inequality}, because there is only the derivative of $h_{s, \alpha}$ about the variable $v$ on the right-hand side of the equality~\eqref{eqn: only-v-inequality}.


\subsection{Proof of Proposition~\ref{sec:linear-convergence}}             
\label{subsec: proposition31}                                                                  
To better appreciate the linear convergence of the hp-dependent SDE~\eqref{eqn: lr-momentum-sde}, as established in Proposition~\ref{sec:linear-convergence}, we need to show  the linear convergence for $H^1$-norm instead of $L^2$-norm,  
which is the key difference with the lr-dependent SDE~\eqref{eqn: lr-sde} in~\cite{shi2020learning}. The technique is named hypocoercivity, introduced by Villani in~\cite{villani2009hypocoercivity}.

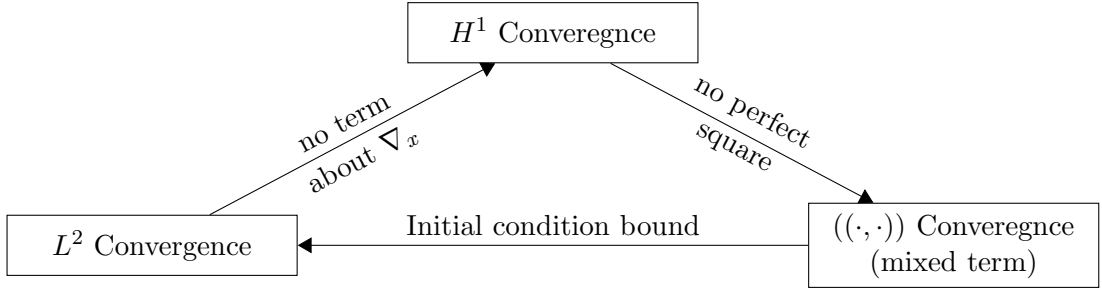
\begin{figure}[!hbtp]
\begin{center}
\begin{tikzpicture}[node distance=4cm]
    \node (n00) [box, draw=black] {$L^{2}$ Convergence };
    \node (n01) [box, draw=black,  above right of=n00, xshift=+2.5cm] {$H^1$ Converegnce};
    \node (n10) [box, draw=black,  below right of=n01, xshift=+2.5cm] {$((\cdot, \cdot))$ Converegnce (mixed term)};

    \draw [arrow] (n00) --  node[sloped, anchor=center, above]{no term} node[sloped, anchor=center, below]{about $\nabla_x$}(n01);
    \draw [arrow] (n01) --  node[sloped, anchor=center, above]{no perfect} node[sloped, anchor=center,below]{square}(n10);
    \draw [arrow] (n10) to node[sloped, anchor=center, above]{Initial condition bound}(n00);
\end{tikzpicture}
\end{center}
\caption{The Diagram of the proof framework for $L^2$-convergence.}
\label{fig:chart}
\end{figure}

\subsubsection{$H^1$-Norm}
\label{subsubsec: h-1-norm}
Due to the failure of Poincar\'e inequality, we need to introduce a new Hilbert space $H^{1}(\mu_{s, \alpha})$, where the inner product is
\[
\left\langle g_{1}, g_{2} \right\rangle_{H^{1}(\mu_{s, \alpha})} =\left\langle g_{1}, g_{2} \right\rangle_{L^{2}(\mu_{s, \alpha})} +  \left\langle \mathscr{A}_{s, \alpha}g_{1}, \mathscr{A}_{s, \alpha}g_{2} \right\rangle_{L^{2}(\mu_{s, \alpha})} +  \left\langle \mathscr{C}_{s, \alpha}g_{1}, \mathscr{C}_{s, \alpha}g_{2} \right\rangle_{L^{2}(\mu_{s, \alpha})} 
\] 
for any $g_1, g_2 \in H^{1}(\mu_{s, \alpha})$. Then, the induced $H^1$-norm for $h_{s, \alpha} - 1$ is 
\begin{align*}
\| h_{s, \alpha} - 1\|_{H^{1}(\mu_{s, \alpha})}^{2}  = \| h_{s, \alpha} - 1\|_{L^{2}(\mu_{s, \alpha})} ^2 + \| \mathscr{A}_{s, \alpha} h_{s, \alpha} \|^2_{L^{2}(\mu_{s, \alpha})}  + \| \mathscr{C}_{s, \alpha} h_{s, \alpha} \|^2_{L^{2}(\mu_{s, \alpha})}.
\end{align*}
\noindent For convenience, we use $h_{s, \alpha}$ instead of $h_{s, \alpha} - 1$. Then, we can obtain the derivative as
\begin{align}
\frac{1}{2} \frac{d}{d t} \left\langle h_{s, \alpha}, h_{s, \alpha}  \right\rangle_{H^{1}(\mu_{s, \alpha})} 
                                                                                                                                               = & (1 + 2\sqrt{\mu}) \left\| \mathscr{A}_{s, \alpha} h_{s, \alpha} \right\|^2_{L^{2}(\mu_{s, \alpha})} +  \| \mathscr{A}_{s, \alpha}^2 h_{s, \alpha} \|^2_{L^{2}(\mu_{s, \alpha})}  
                                                                                                                                               \nonumber \\
                                                                                                                                                  &  + \left\langle \mathscr{C}_{s, \alpha} h_{s, \alpha}, \mathscr{A}_{s, \alpha} h_{s, \alpha}  \right\rangle_{L^{2}(\mu_{s, \alpha})}  + \| \mathscr{A}_{s, \alpha} \mathscr{C}_{s, \alpha}   h_{s, \alpha} \|^2_{L^{2}(\mu_{s, \alpha})} \nonumber \\ 
                                                                                                                                                  & - \left\langle \nabla^2_x f \cdot \mathscr{A}_{s, \alpha} h_{s, \alpha}, \mathscr{C}_{s, \alpha} h_{s, \alpha}  \right\rangle_{L^{2}(\mu_{s, \alpha})}, \label{eqn: derivative-h-s}
\end{align}
where the detailed calculation is shown in Appendix~\ref{subsec: technical-details-h1}. From~\eqref{eqn: derivative-h-s}, we can find only the last term on the right-hand side is negative. The simple Cauchy-Schwartz inequality tells us that there are two terms needed to be bounded as
\begin{equation}
\label{eqn: cauchy-schwartz-inq}
 - \left\langle \nabla^2_x f \cdot \mathscr{A}_{s, \alpha} h_{s, \alpha}, \mathscr{C}_{s, \alpha} h_{s, \alpha}  \right\rangle_{L^{2}(\mu_{s, \alpha})} \geq - \frac12\|\nabla^2_x f \cdot \mathscr{A}_{s, \alpha} h_{s, \alpha}\|_{L^{2}(\mu_{s, \alpha})}^{2} - \frac12\| \mathscr{C}_{s, \alpha} h_{s, \alpha}  \|_{L^{2}(\mu_{s, \alpha})}^{2} 
\end{equation}
The first part needs a bound for $\nabla_x^2 f$, which requires us to consider the Villani condition-(II), shown in Section~\ref{subsubsec: relative-bound}. The second part requires us to consider a mixed term, shown in Section~\ref{subsubsec: mixed-term}.


\subsubsection{Relative bound}
\label{subsubsec: relative-bound}
With the introduction of Villani condition-(II), we can obtain a relative bound as
\begin{lem}[Lemma A.24 in~\cite{villani2009hypocoercivity}]
\label{lem: relative-bound}
\; Let $f \in C^2(\mathbb{R}^d)$ and satisfy the Villani condition-$\text{(II)}$
\[
\left\| \nabla^2 f \right\|_2  \leq C\left( 1 + \left \| \nabla f \right\| \right),
\]
Then, for all $g  \in H^{1}(\mu_{s, \alpha})$, we have
\begin{itemize}
\item[\rm{(i)}] $\| (\nabla_x f) g \|_{L^2(\mu_{s, \alpha})}^{2} \leq  \kappa_1(s, \alpha) \left( \| g \|_{L^2(\mu_{s, \alpha})}^{2} + \| \nabla_x g \|_{L^2(\mu_{s, \alpha})}^{2} \right)$;
\item[\rm{(ii)}] $\| \| \nabla^2_x f \|_2 g \|_{L^2(\mu_{s, \alpha})}^{2} \leq \kappa_2(s, \alpha) \left( \|g\|_{L^2(\mu_{s, \alpha})}^{2} + \| \nabla_x g \|_{L^2(\mu_{s, \alpha})}^{2}  \right)$.
\end{itemize}

\end{lem}
\noindent The proof is shown in~\Cref{subsec: proof-lemma45}.                                               
%
Then, with Lemma~\ref{lem: relative-bound}, we can bound the first term on the right-hand side of~\eqref{eqn: cauchy-schwartz-inq} as
\begin{align*}
                 \left\|  \nabla^2_x f \cdot \mathscr{A}_{s, \alpha} h_{s, \alpha} \right\|_{L^2(\mu_{s, \alpha}) }^2   & =   \| \left[\mathscr{T}, \mathscr{C}_{s, \alpha}\right] h_{s, \alpha} \|_{L^2(\mu_{s, \alpha})}^2  \\& =      \frac{\sqrt{s}}{2} \| \nabla_x^2 f \cdot \nabla_v h_{s, \alpha} \|_{L^2(\mu_{s, \alpha})}^2 \\
                                                                                                                      & \leq  \frac{\kappa_2(s, \alpha) \sqrt{s}}{2}  \bigg( \|\nabla_x\nabla_v h_{s, \alpha} \|_{L^{2}(\mu_{s, \alpha})}^{2} + \|\nabla_v h_{s, \alpha} \|_{L^{2}(\mu_{s, \alpha})}^{2}   \bigg) \\
                                                                                                                      & \leq   \frac{2 \kappa_2(s, \alpha)}{\sqrt{s}} \| \mathscr{A}_{s, \alpha} \mathscr{C}_{s, \alpha} h_{s, \alpha} \|_{L^{2}(\mu_{s, \alpha})}^2 + \kappa_2(s, \alpha) \|\mathscr{A}_{s, \alpha} h_{s, \alpha} \|_{L^{2}(\mu_{s, \alpha})}^2 \\
                                                                                                                      & \leq  \kappa_3(s, \alpha) \bigg(  \| \mathscr{A}_{s, \alpha} \mathscr{C}_{s, \alpha} h_{s, \alpha} \|_{L^{2}(\mu_{s, \alpha})}^{2} +  \|\mathscr{A}_{s, \alpha} h_{s, \alpha} \|_{L^{2}(\mu_{s, \alpha})}^{2}   \bigg),       
 \end{align*}                                       
where $\kappa_3(s, \alpha) = \max\{  2\kappa_2(s, \alpha)/\sqrt{s}, \kappa_2(s, \alpha) \}$.

\subsubsection{Mixed term}
\label{subsubsec: mixed-term}

For the second term on the right-hand side of~\eqref{eqn: cauchy-schwartz-inq}, $-  \| \mathscr{C}_{s, \alpha} h_{s, \alpha}  \|_{L^2(\mu_{s, \alpha})}^2$, in order to make it a perfect sum of squares, we need to consider the mixed term \[\left\langle \mathscr{A}_{s, \alpha} h_{s, \alpha}, \mathscr{C}_{s, \alpha} h_{s, \alpha} \right\rangle_{L^2(\mu_{s, \alpha})},\] of which the derivative is 
\begin{align}
 \frac{d}{d t} \langle \mathscr{A}_{s, \alpha} h_{s, \alpha}, &\mathscr{C}_{s, \alpha} h_{s, \alpha} \rangle_{L^2(\mu_{s, \alpha})} \nonumber \\  \geq  &- 2 \| \mathscr{A}^2_{s, \alpha}  h_{s, \alpha} \|_{L^2(\mu_{s, \alpha})} \| \mathscr{A}_{s, \alpha} \mathscr{C}_{s, \alpha} h_{s, \alpha} \|_{L^2(\mu_{s, \alpha})} \nonumber \\ 
        &- 2\sqrt{\mu} \| \mathscr{A}_{s, \alpha}   h_{s, \alpha} \|_{L^2(\mu_{s, \alpha})} \|   \mathscr{C}_{s, \alpha}  h_{s, \alpha}  \|_{L^2(\mu_{s, \alpha})}  +  \| \mathscr{C}_{s, \alpha}   h_{s, \alpha} \|^2_{L^2(\mu_{s, \alpha})} \nonumber \\
        &- \sqrt{\kappa_3(s, \alpha)} \| \mathscr{A}_{s, \alpha}   h_{s, \alpha}\|_{L^2(\mu_{s, \alpha})} \left( \|\mathscr{A}_{s, \alpha} \mathscr{C}_{s, \alpha} h_{s, \alpha}  \|_{L^2(\mu_{s, \alpha})} +  \| \mathscr{A}_{s, \alpha} h_{s, \alpha}  \|_{L^2(\mu_{s, \alpha})}  \right). \label{eqn: derivative-mixed}
\end{align}
The detailed calculation is shown in Appendix~\ref{subsec: technical-detail-mixed}.


\subsubsection{New inner product}
\label{subsubsec: new-inner-product}
From~\eqref{eqn: derivative-h-s} and~\eqref{eqn: derivative-mixed}, we have obtained the four following squared terms.
\[
\left\| \mathscr{A}_{s, \alpha} h_{s, \alpha} \right\|^2_{L^2(\mu_{s, \alpha})}, \quad \| \mathscr{A}^2_{s, \alpha} h_{s, \alpha} \|^2_{L^2(\mu_{s, \alpha})}, \quad  \| \mathscr{A}_{s, \alpha} \mathscr{C}_{s, \alpha}   h_{s, \alpha} \|^2_{L^2(\mu_{s, \alpha})}, \quad \| \mathscr{C}_{s, \alpha}   h_{s, \alpha} \|^2_{L^2(\mu_{s, \alpha})}.
\]
Intuitively, we can construct a new inner product satisfying 
\[
- \frac{1}{2} \frac{d}{dt} \left(( h_{s, \alpha}, h_{s, \alpha} \right)) \geq \lambda(s, \alpha) \left(( h_{s, \alpha}, h_{s, \alpha} \right)), 
\]
where $\lambda(s, \alpha)$ is some positive constant which depends on the parameters $s$ and $\alpha$. If $b^2 \leq ac$, we can define the new inner product as
\begin{align}
\left(\left( h_{s, \alpha}, h_{s, \alpha} \right)\right) =   \left\langle h_{s, \alpha}, h_{s, \alpha} \right\rangle&_{L^2(\mu_{s, \alpha})}   +  a\left\langle \mathscr{A}_{s, \alpha}h_{s, \alpha}, \mathscr{A}_{s, \alpha} h_{s, \alpha} \right\rangle_{L^2(\mu_{s, \alpha})} \nonumber \\ &+ 2b  \left\langle \mathscr{A}_{s, \alpha} h_{s, \alpha}, \mathscr{C}_{s, \alpha} h_{s, \alpha} \right\rangle_{L^2(\mu_{s, \alpha})} + c \left\langle \mathscr{C}_{s, \alpha}h_{s, \alpha}, \mathscr{C}_{s, \alpha}h_{s, \alpha} \right\rangle_{L^2(\mu_{s, \alpha})} \label{eqn: new-inner-product}
\end{align}
Apparently, the norm $\| \cdot \|_{((\cdot, \cdot))}$ induced by the new inner product $((\cdot, \cdot))$ is equivalent to $H^1$-norm, that is, there exists two positive reals $\mathcal{C}_1$ and $\mathcal{C}_2$ such that
\begin{equation}
\label{eqn: norm-equivalent}
\mathcal{C}_1\| h_{s, \alpha}\|_{((\cdot, \cdot))}^2 \leq  \| h_{s, \alpha}\|_{H^1}^2 \leq \mathcal{C}_2\| h_{s, \alpha}\|_{((\cdot, \cdot))}^2. 
\end{equation}
 \noindent Then, from the basic calculations in Section~\ref{subsubsec: h-1-norm} and Section~\ref{subsubsec: mixed-term}, we can obtain the following expression by combining like terms as
\begin{align}
- \frac{1}{2} \frac{d}{dt} \left(\left( h_{s, \alpha}, h_{s, \alpha} \right)\right)  & =  \left(\left( \mathscr{L}_{s, \alpha}h_{s, \alpha}, h_{s, \alpha} \right)\right)  \nonumber \\
                                                                                                         & \geq \left(1 + 2a\sqrt{\mu} -2b\sqrt{\kappa_3(s, \alpha)} \right)  \left\| \mathscr{A}_{s, \alpha} h_{s, \alpha} \right\|^2_{L^2(\mu_{s, \alpha})} + a \| \mathscr{A}^2_{s, \alpha}  h_{s, \alpha} \|^2_{L^2(\mu_{s, \alpha})} \nonumber \\
                                                                                                         & \mathrel{\phantom{=}}  + c  \| \mathscr{A}_{s, \alpha}  \mathscr{C}_{s, \alpha}    h_{s, \alpha}  \|^2 _{L^2(\mu_{s, \alpha})}+ 2b \| \mathscr{C}_{s, \alpha}    h_{s, \alpha}  \|^2_{L^2(\mu_{s, \alpha})} \nonumber \\
                                                                                                         & \mathrel{\phantom{=}} - 2b\sqrt{\kappa_3(s, \alpha)}  \| \mathscr{A}_{s, \alpha}    h_{s, \alpha}  \| _{L^2(\mu_{s, \alpha})}  \| \mathscr{A}_{s, \alpha}  \mathscr{C}_{s, \alpha}    h_{s, \alpha} \| _{L^2(\mu_{s, \alpha})} \nonumber \\
                                                                                                         &\mathrel{\phantom{=}}  - \left( a + c\sqrt{\kappa_3(s, \alpha)}+ 4b\sqrt{\mu}\right)  \| \mathscr{A}_{s, \alpha}    h_{s, \alpha}  \|_{L^2(\mu_{s, \alpha})}   \| \mathscr{C}_{s, \alpha}   h_{s, \alpha} \|_{L^2(\mu_{s, \alpha})} \nonumber \\
                                                                                                         & \mathrel{\phantom{=}} - 4b \| \mathscr{A}_{s, \alpha}^2  h_{s, \alpha} \|_{L^2(\mu_{s, \alpha})} \| \mathscr{A}_{s, \alpha} \mathscr{C}_{s, \alpha} h_{s, \alpha} \|_{L^2(\mu_{s, \alpha})} \nonumber \\
                                                                                                         & \mathrel{\phantom{=}} -c\sqrt{\kappa_3(s, \alpha)}  \left\| \mathscr{A}_{s, \alpha} \mathscr{C}_{s, \alpha}  h_{s, \alpha} \right\|_{L^2(\mu_{s, \alpha})}  \left\| \mathscr{C}_{s, \alpha}  h_{s, \alpha} \right\|_{L^2(\mu_{s, \alpha})} \label{eqn: deriative-for-mixed}
\end{align}
From~\eqref{eqn: deriative-for-mixed}, we can find this problem is transformed to guarantee the following matrix $K_1$ positive definite as
\[
K_1 = 
\begin{pmatrix}
1 + 2a\sqrt{\mu} -2b\sqrt{\kappa_3(s, \alpha)}               & 0                         &  -b\sqrt{\kappa_3(s, \alpha)}           &  -\frac{1}{2}  \left(a + c\sqrt{\kappa_3(s, \alpha)}+ 4b\sqrt{\mu}\right)    \\
               0                                                                        &  a                      &  -2b                                               &      0                                                         
             \\
           -b\sqrt{\kappa_3(s, \alpha)}                                   &  -2b                   &  c                                                 & -\frac{1}{2}c\sqrt{\kappa_3(s, \alpha)}                           
              \\
-\frac{1}{2}  \left(a + c\sqrt{\kappa_3(s, \alpha)}+ 4b\sqrt{\mu}\right)          &   0                       &  -\frac{1}{2}c\sqrt{\kappa_3(s, \alpha)}                    & 2b
\end{pmatrix}.
\]
Let $M: = \max\{1, \sqrt{\mu}, \sqrt{\kappa_3(s, \alpha)}\}$ and fix $1 \geq a \geq b \geq 2c$, in order to make the matrix $K$ positive definite, we can make  the following matrix $K_2$ positive definite as
\[
K_2 = \begin{pmatrix}
1 - 2Ma        &  0         & -Mb                    &  -3Ma                  \\
0                  &  a         & -2Mb                  &      0                    \\
-Mb              &-2Mb     & c                        &  - \frac{1}{2}Mc   \\
-3Ma            &  0         & - \frac{1}{2}Mc   &  2b
\end{pmatrix}.
\]
Furthermore, if we  assume that  $M \leq 1/4a$, we here use the matrix $L$ instead of $K_2$ as
\[
L =  [l_{ij}] \equiv\begin{pmatrix}
\frac{1}{2}     &  0         & -Mb                    &  -3Ma                  \\
0                  &  a         & -2Mb                  &      0                    \\
-Mb              &-2Mb     & c                        &  - \frac{1}{2}Mc   \\
-3Ma            &  0         & - \frac{1}{2}Mc   &  2b
\end{pmatrix}. 
\]

Recall the fact that if the element $l_{ij}$ of the matrix $L$ satisfies that 
\begin{equation}
\label{eqn: positive-definite-condition}
|l_{ij}| = |l_{ji}| \leq \frac{ \sqrt{ l_{ii} l_{jj} } }{4} \leq \frac{l_{ii} + l_{jj}}{8},
\end{equation}
the following quadratic inequality will be obtained as 
\begin{align*}
\sum_{i=1}^4 \sum_{j=1}^4 l_{ij}x_ix_j \geq  \frac{1}{4} \sum_{i=1}^{d}l_{ii} x_i^2.
\end{align*}
To ensure~\eqref{eqn: positive-definite-condition}, we require the coefficient $M$ satisfies that 
\[
Mb \leq \frac{\sqrt{c/2}}{4}, \qquad 2Mb \leq \frac{\sqrt{ac}}{4}, \qquad 3Ma \leq \frac{\sqrt{b}}{4}, \qquad \frac{1}{2}Mc \leq \frac{\sqrt{2bc}}{4},
\]
that is,
\[
M =  \sqrt{\min\left\{ 1, \; \frac{1}{4a},\; \frac{c}{32b^2}, \; \frac{ac}{64b^2}, \; \frac{b}{144a^2}, \; \frac{b}{2c} \right\}}.
\]
Then, we can obtain the following estimate for~\eqref{eqn: deriative-for-mixed} as
\begin{align*}
- \frac{1}{2} \frac{d}{dt} \left(\left( h_{s, \alpha}, h_{s, \alpha} \right)\right) &     =  \left(\left( \mathscr{L}_{s, \alpha} h_{s, \alpha}, h_{s, \alpha} \right)\right) \\
                                                                                                         & \geq \frac{1}{4} \min\left\{ \frac{1}{2}, 2b \right\} \left(  \left\| \mathscr{A}_{s, \alpha} h_{s, \alpha} \right\|^2_{L^2(\mu_{s, \alpha})}  +  \left\| \mathscr{C}_{s, \alpha} h_{s, \alpha} \right\|^2_{L^2(\mu_{s, \alpha})}  \right) \\
                                                                                                         & \geq   \frac{1}{8} \min\left\{ \frac{1}{2}, 2b \right\} \left(  \left\| \mathscr{A}_{s, \alpha} h_{s, \alpha} \right\|^2_{L^2(\mu_{s, \alpha})}  +  \left\| \mathscr{C}_{s, \alpha} h_{s, \alpha} \right\|^2_{L^2(\mu_{s, \alpha})}  \right) \\
                                                                                                         & \mathrel{\phantom{\geq}}+  \frac{1}{8} \min\left\{ \frac{1}{2}, 2b \right\} \chi_{ s, \alpha} \| h_{s, \alpha}\|^2_{L^2(\mu_{s, \alpha})} \\
                                                                                                         & \geq \min\left\{\frac{1}{8} \min\left\{ \frac{1}{2}, 2b \right\},  \frac{1}{8} \min\left\{ \frac{1}{2}, 2b \right\} \chi_{ s, \alpha} \right\} \| h_{s, \alpha}\|^2_{H^1(\mu_{s, \alpha})}
\end{align*}
where the last but one inequality follows Poincar\'e inequality~\eqref{eqn: poincare}. With the norm equivalence~\eqref{eqn: norm-equivalent} and taking 
\[
\lambda_{s, \alpha} = \mathcal{C}_1 \cdot \min\left\{\frac{1}{8} \min\left\{ \frac{1}{2}, 2b \right\},  \frac{1}{8} \min\left\{ \frac{1}{2}, 2b \right\} \chi_{ s, \alpha} \right\},
\]
we can furthermore obtain the following inequality as
\[
- \frac{1}{2} \frac{d}{dt} \left(\left( h_{s, \alpha}, h_{s, \alpha} \right)\right)  \geq \lambda_{s, \alpha}   \left(\left( h_{s, \alpha}, h_{s, \alpha}\right)\right). 
\]
Continuing with the norm equivalence~\eqref{eqn: norm-equivalent}, we have
\[
\|h_{s, \alpha}(t, \cdot, \cdot)\|_{L^{2}(\mu_{s, \alpha})}^{2} \leq e^{-2 \lambda_{s, \alpha} t}   \left(\left( h_{s, \alpha}(0), h_{s, \alpha}(0) \right)\right) \leq \mathcal{C}_{2} e^{-2 \lambda_{s, \alpha} t}  \|h_{s, \alpha}(0, \cdot, \cdot)\|_{H^{1}(\mu_{s, \alpha})}^{2}. 
\]
%
Finally, from~\cite[Theorem A.8]{villani2009hypocoercivity}, we know that the $H^1$-norm $\| h_{s, \alpha}(t, \cdot, \cdot)\|_{H^1(\mu_{s, \alpha})}$ at $0<t \leq 1$ can be bounded by $L^2$-norm $\| h_{s, \alpha}(0, \cdot, \cdot) \|_{L^2(\mu_{s, \alpha})}$ at $t = 0$. Some basic operations tell us that there exists $\mathcal{C}_3 > 0$ such that
\[
\left\|\rho_{s, \alpha}(t, \cdot, \cdot) - \mu_{s, \alpha}\right\|_{L^{2}(\mu_{s, \alpha}^{-1})}^{2} \leq  \mathcal{C}_3e^{-2 \lambda_{s, \alpha} t} \left \|\rho_{s, \alpha}(0, \cdot, \cdot) - \mu_{s, \alpha} \right\|_{L^{2}(\mu_{s, \alpha}^{-1})}^{2}.
\]

\subsection{Proof of Proposition~\ref{prop: final-gap}}
\label{subsec: proof-final-gap}
Since the potential, $f(x) - f^\star$, is only for $x$ in the proof of Proposition~\ref{prop: final-gap}, we can integrate the variable $v$. Hence, except for the mixing parameter $\beta(s, \alpha)$ instead of the learning rate $s$, the technique here is the same as that used in~\cite{shi2020learning}.

%% file: quantitative.tex
\section{Estimate of Exponential Decay Constant}
\label{sec: quantative}

In this section, we complete the proof of Theorem~\ref{thm: continuous-quantative} to quantify the linear rate of linear convergence, the exponential decay constant $\lambda_{s, \alpha}$. This is crucial for us to understand the dynamics of SGD with momentum, especially its dependence on the hyperparameters, the learning rate $s$ and the momentum coefficient $\alpha$.

\subsection{Connection to the Kramers-Fokker-Planck operator}                                         
\label{subsec: connect-KFP-operators}                                                         

Similar as~\cite{shi2020learning}, we start to derive the relationship between the hp-dependent SDE~\eqref{eqn: lr-momentum-sde-standard} and the Kramers operator. Recall that the probability density $\rho_{s, \alpha}(t, \cdot, \cdot)$ of the solution to the hp-dependent SDE~\eqref{eqn: lr-momentum-sde-standard} is assumed in $L^2(\mu_{s, \alpha}^{-1})$. Here, we consider the transformation
\[
\psi_{s, \alpha}(t, \cdot, \cdot) = \frac{\rho_{s, \alpha}(t, \cdot, \cdot)}{\sqrt{\mu_{s, \alpha}}} \in L^{2}(\mathbb{R}^d, \mathbb{R}^d).
\]
With this transformation, we can equivalently rewrite the kinetic Fokker-Planck equation~\eqref{eqn: kinetic-FP} as
\begin{equation}
\label{eqn: FP-Kramers}
\frac{\partial \psi_{s, \alpha}}{\partial t} = - \left[ v \cdot  \nabla_x - \nabla_x f \cdot \nabla_v -  \frac{\sqrt{s}}{2}  \left( \Delta_v + \frac{d}{\beta(s, \alpha)}- \frac{\|v\|^2}{\beta^2(s, \alpha)} \right) \right]
\end{equation}
with the initial $\psi_{s, \alpha}(0, \cdot, \cdot) = \rho_{s, \alpha}(0, \cdot, \cdot)/ \sqrt{\mu_{s, \alpha}} \in L^{2}(\mathbb{R}^d, \mathbb{R}^d)$. Here, the operator on the right-hand side of the kinetic Fokker-Planck equation~\eqref{eqn: FP-Kramers} is named as Kramers operator:
\begin{equation}
\label{eqn: kramers-operator}
\mathscr{K}_{s, \alpha} = v \cdot  \nabla_x - \nabla_x f \cdot \nabla_v -  \frac{\sqrt{s}}{2}  \left( \Delta_v + \frac{d}{\beta(s, \alpha)}- \frac{\|v\|^2}{\beta^2(s, \alpha)} \right).
\end{equation}

Similarly, we collect some essential facts concerning the spectrum of the Kramers operator $\mathscr{K}_{s, \alpha}$. Here, we first need to show the spectrum of the Kramers operator $\mathscr{K}_{s, \alpha}$ is discrete, that is, the unbounded Kramers operator $\mathscr{K}_{s, \alpha}$ has a compact resolvent. Based on the Villani condition-(I), the unbounded Kramers operator $\mathscr{K}_{s, \alpha}$ has a compact resolvent, first shown in~\cite[Corollary 5.10, Remark 5.13]{helffer2005hypoelliptic}. However, it still needs a polynomial-controlled condition.  Until~\cite{li2012global}, the polynomial-controlled condition has not been removed. Here, we state the well-known result in spectral theory --- that the unbounded Kramers operator $\mathscr{K}_{s, \alpha}$ has a compact resolvent~\cite{li2012global}. 
\begin{lem}[Corollary 1.4 in~\cite{li2012global}]
\label{lem: discrete-spectral}
Let the potential $f(x)$ satisfy the Villani conditions (Definition~\ref{defn: villani}). The Kramers operator $\mathscr{K}_{s, \alpha}$ has a compact resolvent.
\end{lem}

Recall the existence and uniqueness of Gibbs distribution $\mu_{s, \alpha}$ in~\Cref{subsec: exist-unique-gibbs}, it is not hard to show that $\sqrt{\mu_{s, \alpha}}$ is the unique eigenfunction of  $\mathscr{K}_{s, \alpha}$ corresponding to the eigenvalue zero. Taking~\eqref{eqn: only-v-inequality}, we can obtain 
\begin{align*}
\left\langle \mathscr{K}_{s, \alpha}\psi_{s, \alpha}(t, \cdot, \cdot),  \psi_{s, \alpha}(t, \cdot, \cdot) \right\rangle_{L^2(\mathbb{R}^d, \mathbb{R}^d)} & = \left\langle \mathscr{K}_{s, \alpha}(\psi_{s, \alpha}(t, \cdot, \cdot) - \sqrt{\mu_{s, \alpha}}), \psi_{s, \alpha}(t, \cdot, \cdot) - \sqrt{\mu_{s, \alpha}} \right\rangle_{L^2(\mathbb{R}^d, \mathbb{R}^d)} \\
                                                                                                                                                              & = - \frac{1}{2}\frac{d}{dt}\left\langle \psi_{s, \alpha}(t, \cdot, \cdot) - \sqrt{\mu_{s, \alpha}}, \psi_{s, \alpha}(t, \cdot, \cdot) - \sqrt{\mu_{s, \alpha}} \right\rangle_{L^2(\mathbb{R}^d, \mathbb{R}^d)} \\
                                                                                                                                                              & = - \frac{1}{2}\frac{d}{dt} \left\| \rho_{s, \alpha}(t, \cdot, \cdot) - \mu_{s, \alpha} \right\|_{L^2(\mu_{s, \alpha})}^{2} \geq 0.
\end{align*}
With Lemma~\ref{lem: discrete-spectral}, this verifies the unbounded Kramers operator $\mathscr{K}_{s, \alpha}$ is positive semidefinite.  Hence, we can order the eigenvalues of $\mathscr{K}_{s, \alpha}$ in $L^2(\mathbb{R}^d, \mathbb{R}^d)$ as
\[
0 = \zeta_{s, \alpha; 0} \leq \zeta_{s, \alpha; 1} \leq \cdots \leq \zeta_{s, \alpha; \ell} \leq \cdots \leq + \infty.
\]
Recall Theorem~\ref{thm: continuous-qualitative}, the exponential decay constant in~\eqref{eqn: continuous-qualitative} is set to 
\[
\lambda_{s, \alpha} = \zeta_{s, \alpha; 1}.
\]
To see this, note that $\psi_{s, \alpha}(t, \cdot, \cdot) - \sqrt{\mu_{s, \alpha}}$ also satisfies~\eqref{eqn: FP-Kramers} and is orthogonal to the null eigenfunction $\sqrt{\mu_{s, \alpha}}$. Therefore, we can obtain
\begin{align*}
\left\langle \mathscr{K}_{s, \alpha}(\psi_{s, \alpha}(t, \cdot, \cdot) - \sqrt{\mu_{s, \alpha}}), \psi_{s, \alpha}(t, \cdot, \cdot) - \sqrt{\mu_{s, \alpha}} \right\rangle_{L^2(\mathbb{R}^d, \mathbb{R}^d)} & \geq \zeta_{s, \alpha; 1} \| \psi_{s, \alpha}(0, \cdot, \cdot) - \sqrt{\mu_{s, \alpha}}) \|_{L^{2}(\mathbb{R}^d, \mathbb{R}^d)}^2 \\
                                                                               & = \zeta_{s, \alpha; 1} \| \rho_{s, \alpha}(0, \cdot, \cdot) - \mu_{s, \alpha} \|_{L^{2}(\mu_{s, \alpha}^{-1})}^{2}.
\end{align*}
Furthermore, we can obtain the exponential estimate for $\rho_{s, \alpha}(t, \cdot, \cdot) - \mu_{s, \alpha}$ as
\[
\| \rho_{s, \alpha}(t, \cdot, \cdot) - \mu_{s, \alpha} \|_{L^{2}(\mu_{s, \alpha}^{-1})}^{2} \leq e^{- \zeta_{s, \alpha; 1}t} \| \rho_{s, \alpha}(0, \cdot, \cdot) - \mu_{s, \alpha} \|_{L^{2}(\mu_{s, \alpha}^{-1})}^{2}.
\]

\subsection{The spectrum of Kramers-Fokker-Planck operators}                
\label{subsec: spectrum-KFP-operators}                                                        
Similar in~\cite{shi2020learning}, the Schr\"odinger operator is equivalent to the Witten-Laplacian, here the unbounded Kramers operator $\mathscr{K}_{s, \alpha}$ is equivalent to the Kramers-Fokker-Planck operator as
\begin{equation}
\label{eqn: Kramers-FP}
\mathscr{P}_{s, \alpha} := \beta(s, \alpha)\mathscr{K}_{s, \alpha} = v \cdot \beta(s, \alpha) \nabla_x - \nabla_x f \cdot \beta(s, \alpha) \nabla_v + \sqrt{\mu} (v - \beta(s, \alpha) \nabla_v)(v + \beta(s, \alpha) \nabla_v),
\end{equation}
where $\mathscr{P}_{s, \alpha}$ is only a simple rescaling of $\mathscr{K}_{s, \alpha}$. Denote the eigenvalues of the Kramers-Fokker-Planck operator as $0 = \delta_{s, \alpha; 0} \leq \delta_{s, \alpha; 1} \leq \cdots \leq \delta_{s, \alpha; \ell} \leq \cdots \leq +\infty$, we get the simple relationship as
\[
\delta_{s, \alpha; \ell} = \beta(s, \alpha) \zeta_{s, \alpha; \ell}
\]
for all $\ell \in \mathbb{N}$. 

The spectrum of the Kramers-Fokker-Planck operator $\mathscr{P}_{s, \alpha}$ has been investigated in~\cite{helffer2005hypoelliptic, herau2011tunnel}. Here, we exploit this literature to derive the closed-form expression for the first positive eigenvalue of the Kramers-Fokker-Planck operator,  thereby obtaining the dependence of the exponential decay constant on the hyperparameters, the learning rate $s$ and the momentum coefficient $\alpha$, for a certain class of nonconvex objective functions.

Recall in Section~\ref{sec: prelim}, and we show the basic concepts of the Morse function. For detail about the ideas of the Morse function, please refer to~\cite[Section 6.2]{shi2020learning}. Notably, the most crucial concept, the index-$1$ separating saddle (See~\cite[Figure 9]{shi2020learning}), is introduced. Here, we briefly describe the general assumption for a Morse function.  

\begin{assumption}[Generic case \cite{herau2011tunnel}, also see Assumption 6.4~\cite{shi2020learning}]\label{assump: generic-assumption}
For every critical component $E^{i}_j$ selected in the labeling process above, where $i = 0, 1, \ldots, I$,  we assume that
\begin{itemize}
\item The minimum $x^\bullet_{i,j}$ of $f$ in any critical component $E^{i}_j$ is unique.
\item If $E_{j}^{i} \cap \mathcal{X}^\circ \neq \varnothing$, there exists a unique $x_{i,j}^{\circ} \in E_{j}^{i} \cap \mathcal{X}^\circ$ such that $f(x_{i,j}^{\circ}) = \max\limits_{x \in E_{j}^{i} \cap \mathcal{X}^\circ} f(x)$. In particular, $E_j^i \cap \mathcal{K}_{f(x_{i,j}^{\circ})}$ is the union of two distinct critical components. 

\end{itemize}
\end{assumption}

With the generic assumption (Assumption~\ref{assump: generic-assumption}), the labeling process for the index-1 separating saddle and local minima (See~\cite[Figure 10]{shi2020learning}) is introduced, which reveals a remarkable result: there exists a bijection between the set of local minima and the set of index-$1$ separating saddle points (including the fictive one) $\mathcal{X}^{\circ} \cup \{\infty\}$. Interestingly, this shows that the number of local minima is always larger than the number of index-1 separating saddle points by one; that is, $n^\circ = n^\bullet - 1$. Here, we relabel the index-1 separating saddle points $x^\circ_\ell$ for $\ell = 0, 1, \ldots, n^{\circ}$ with $x^\circ_0 = \infty$, and the local minima $x^\bullet_\ell$ for $\ell = 0, 1, \ldots, n^{\bullet}-1$ with $x^\bullet_0 = x^\star$, such that
\begin{equation}
\label{eqn: saddle-minima-pair}
f(x^\circ_0) - f(x^\bullet_0) > f(x^\circ_1) - f(x^\bullet_1) \ge \ldots \ge f(x^\circ_{n^{\bullet}-1}) - f(x^\bullet_{n^{\bullet}-1}),
\end{equation}
where $f(x^\circ_0) - f(x^\bullet_0) = f(\infty) - f(x^\star) = +\infty$. A detailed description of this bijection is given in~\cite[Proposition 5.2]{herau2011tunnel}. With the pairs $(x^\circ_{\ell}, x^\bullet_{\ell})$ in place, we state the fundamental result concerning the first $n^{\bullet}-1$ smallest positive eigenvalues of the Fokker-Planck-Kramers operator as
\begin{equation}
\label{eqn: fp-kramers-gamma}
\mathscr{P}_{\beta(s, \alpha)} = v \cdot \beta(s, \alpha) \nabla_x - \nabla_x f \cdot \beta(s, \alpha) \nabla_v + \frac{1}{2}\gamma (v - \beta(s, \alpha) \nabla_v)(v + \beta(s, \alpha) \nabla_v).
\end{equation}


\begin{prop}[Theorem 1.2 in~\cite{herau2011tunnel}]\label{thm: HHS-generic}
Under Assumption~\ref{assump: generic-assumption} and the assumptions of Theorem~\ref{thm: continuous-quantative}, there exists $\beta_0 > 0$ such that for any $\beta \in (0, \beta_0]$, the first $n^{\bullet}-1$ smallest positive eigenvalues of the Kramers-Fokker-Planck operator $\mathscr{P}_{s, \alpha}$ associated with $f$ satisfy
\begin{equation}\label{eqn: asymptotic-generic}
\delta_{\beta(s, \alpha),\ell} = \beta(s, \alpha)|\eta_{d}(x_{\ell}^\circ)|\left(\gamma_\ell + o(\beta(s, \alpha)) \right)  e^{- \frac{2 (f(x_{\ell}^\circ) - f(x_\ell^\bullet) )}{\beta(s, \alpha)}}
\end{equation}
for $\ell =1, 1, \ldots,  n^{\bullet}-1$, where
\[
\gamma_\ell = \frac{1}{\pi} \left( \frac{\det( \nabla^{2} f(x_\ell^\bullet) )}{- \det (\nabla^{2} f(x_\ell^\circ) )}\right)^{\frac{1}{2}},
\]
and $\eta_{d}(x_\ell^\circ)$ is the unique negative eigenvalue of the block matrix 
\[
\begin{pmatrix}
0 & \mathbf{I}_{d \times d} \\
-\nabla^{2} f(x_\ell^\circ) & \gamma \mathbf{I}_{d \times d}
\end{pmatrix}.
\]
\end{prop}

Recall Theorem~\ref{thm: continuous-quantative}, we assume the negative eigenvalue of $\nabla^{2} f(x_1^\circ)$ is $-v$.  Apply Proposition~\ref{thm: HHS-generic} into the Kramers operator $\mathscr{K}_{s, \alpha}$ in~\eqref{eqn: Kramers-FP}, we can obtain there exists $H_{f}>0$ completely depending only on $f$ such that
\[
\zeta_{s, \alpha; 1} = |\eta_{d}(x_{1}^\circ)|\left(\gamma_1 + o(\beta(s, \alpha)) \right)  e^{- \frac{2 H_f}{\beta(s, \alpha)}} =|\eta_{d}(x_{1}^\circ)|\left(\gamma_1 + o(s) \right)  e^{- \frac{2 H_f}{s} \cdot \frac{2(1 - \alpha)}{1 + \alpha}},
\]
where $\eta_{d}(x_{1}^\circ) = \sqrt{\mu} - \sqrt{\mu + v}$. With some basic substitution of notations, we complete the proof of Theorem~\ref{thm: continuous-quantative}.

%% file: nesterov.tex
\section{On Nesterov's Momentum with Noise}
\label{sec: stochastic-nesterov}

This section briefly discusses Nesterov's accelerated gradient descent method (NAG) under the gradient with incomplete information. Recall the NAG, and the algorithms are set as

\begin{equation}
\label{eqn: nesterov-stochastic}
\left\{ \begin{aligned}
         & y_{k+1} = x_{k} - s\nabla f(x_k) + s \xi_k \\
         & x_{k+1} = y_{k+1} + \alpha (y_{k+1} - y_{k}).
         \end{aligned}\right.
\end{equation}
Recall the classical analysis for NAGs in~\cite{shi2018understanding}, and there are two kinds of NAGs, NAG-\texttt{SC} and NAG-\texttt{C}.  

\subsection{NAG-\texttt{SC}}
\label{subsec: nag-sc}

For NAG-\texttt{SC}, Nesterov's accelerated gradient descent method for the $\mu$-strongly convex objective, the momentum coefficient is set as            
$\alpha = \frac{1 - \sqrt{\mu s}}{1 + \sqrt{\mu s}}$. Notably, though the name, NAG-\texttt{SC}, comes from the $\mu$-strongly convex case, here, the algorithm works on the non-convex objective. Similarly, plugging the two high-resolution Taylor expansions,~\eqref{eqn: taylor-3} and~\eqref{eqn: noise-continuous}, into the NAG-\texttt{SC}, we then get the high-resolution SDE for NAG-\texttt{SC} as
\begin{equation}
\label{eqn: high-res-sc}
\ddot{X} + \left( 2\sqrt{\mu} + \sqrt{s} \nabla^2f(X) \right) \dot{X} + (1 + \sqrt{\mu s}) \nabla f(X) = \sqrt[4]{s}\dot{W}.
\end{equation}
Here, when the momentum coefficient is set $\alpha = 0.9$ and the learning rate $s$ is small enough, a simple transformation tells us that the main part for the logarithm of exponential decay constant in Theorem~\ref{thm: continuous-quantative} for the NAG-\texttt{SC} high-resolution SDE~\eqref{eqn: high-res-sc} here is 
\[
- \frac{2H_{(1 + \sqrt{\mu s})f}}{\sqrt{s}/\left(2\sqrt{\mu} + \sqrt{s}\nabla^2 f \right) } = - \frac{2}{1 + \alpha} \cdot \left( 1 + \beta \nabla^2 f(x)\right)  \cdot \frac{2H_f}{ \beta(s, \alpha)  } \approx  - \frac{2H_f}{ \beta(s, \alpha)  }.
\]
Meanwhile, taking a simple calculation, when the learning rate $s$ is very small, we can obtain the asymptotic estimate for $ |\eta_{d}(x_{1}^\circ)|$ in~\eqref{eqn: asymptotic-generic} as
\[
 |\eta_{d}(x_{1}^\circ)| = \sqrt{\mu + \frac{sv^2}{4} + v} - \sqrt{\mu} - \frac{\sqrt{s} v}{2} \approx \sqrt{\mu  + v} - \sqrt{\mu}.
\]
Therefore, we find when the noise exists, the quantitive estimate of the exponential decay constant, $\lambda_{s, \alpha}$, in~\eqref{eqn: continuous-qualitative} is almost the same between the high-resolution SDE~\eqref{eqn: high-res-sc} and the low-resolution SDE~\eqref{eqn: lr-momentum-sde-standard} (the hp-dependent SDE) for NAG-\texttt{SC}.

\subsection{NAG-\texttt{C}}
\label{subsec: nag-c}

Similarly, we consider NAG-\texttt{C}, Nesterov's accelerated gradient descent method for the general convex objective, which works on the non-convex objective function. Plugging the two high-resolution Taylor expansions,~\eqref{eqn: taylor-3} and~\eqref{eqn: noise-continuous}, into the NAG-\texttt{C}, we then get the high-resolution SDE for NAG-\texttt{C} as
\begin{equation}
\label{eqn: high-res-c}
\ddot{X} + \left( \frac{3}{t} + \sqrt{s} \nabla^2f(x) \right) \dot{X} + \left(1 + \frac{3\sqrt{s}}{2t} \right) \nabla f(X) =  \sqrt[4]{s}\dot{W}.
\end{equation}
Consider that the learning rate $s$ is small enough. 
When the time $t$ is small enough, that is, $t \rightarrow 0$, the central part for the logarithm of exponential decay constant in Theorem~\ref{thm: continuous-quantative} is
\[
- \frac{2H_{\left(1 + \frac{3\sqrt{s}}{2t} \right) f}}{\sqrt{s}/  \left( \frac{3}{t} + \sqrt{s} \nabla^2f(X) \right) } \approx -\frac{1}{k}  \left(1 + \frac{3}{2k} \right) \cdot \frac{6H_f}{s}.
\]
Here, we can find the convergence speed is faster at the beginning. However, the change is very sharp, thus, the convergence rate decays very fast. Here, we can also find the high-resolution SDE~\eqref{eqn: high-res-c} for NAG-\texttt{C} is not faster than the hp-dependent SDE~\eqref{eqn: lr-momentum-sde-standard}  under the existence of noise. 


%
%
%
%

%% file: conclu.tex
\section{Conclusion}
\label{sec: conclu}

In~\cite{shi2020learning}, the authors show the linear convergence of SGD in non-convex optimization and quantify the linear rate as the exponential of the negatively inverse learning rate by taking its continuous surrogate. This paper presents the theoretical perspective on the linear convergence of SGD with momentum. Different from SGD, the two hyperparameters together, the learning rate $s$ and the momentum coefficient $\alpha$, play a significant role in the linear convergence rate in the non-convex optimization. Taking the hp-dependent SDE, we proceed further to use modern mathematical tools such as hypocoercivity and semiclassical analysis to analyze the dynamics of SGD with momentum in a continuous-time model. We demonstrate how the linear rate of convergence and the final gap for SGD only for the learning rate $s$ varies with the momentum coefficient $\alpha$ when the momentum is added. Finally, we also briefly analyze the Nesterov momentum under the existence of noise, which has no essential difference from the standard momentum.

Similarly, for understanding theoretically the stochastic optimization via SDEs, the pressing question is to characterize better the gap between the stationary distribution of the hp-dependent SDE and that of the discrete SGD with momentum~\cite{pavliotis2014stochastic}. Moreover, Theorem~\ref{thm: main1} is a bound for the algorithms in finite time $T$, which is based on the numerical analysis and not an algorithmic bound.  A related question is whether Theorem~\ref{thm: main1} can be improved to an algorithmic bound 
\[
\E f(x_{k}) - f^\star \leq O(\beta(s, \alpha) + (1 - \lambda_{s, \alpha} \beta(s, \alpha))^{k}) \quad \text{or} \quad O(\beta(s, \alpha) + (1 + \lambda_{s, \alpha} \beta(s, \alpha))^{-k}).
\]

A straightforward direction is to extend our SDE-based analysis to various learning rate schedules used in training deep neural networks, such as the diminishing learning rate, cyclical learning rates, RMSProp, and Adam~\cite{bottou2018optimization,smith2017cyclical, tieleman2012lecture,kingma2014adam}.   
Moreover, perhaps these results can be used to choose the neural network architecture and the loss function to get a small value of the Morse saddle barrier $H_f$. Similarly, the hp-dependent SDE might give insights into the generalization properties of neural networks, such as implicit regularization~\cite{zhang2016understanding,gunasekar2018characterizing}.

%% file: appendix/proof_kin_sde.tex
\section{Technical Details for Section~\ref{sec: prelim}}
\label{sec: proof_sec_prelim}

Here, we rewrite the hp-dependent SDE~\eqref{eqn: lr-momentum-sde-standard} in the form of a vector field as
\[
d \begin{pmatrix}
    X \\
    V
   \end{pmatrix} = \begin{pmatrix}
   V\\
     - 2\sqrt{\mu} V - \nabla f(X) 
   \end{pmatrix} d t +  \begin{pmatrix}
   0\\
   \sqrt[4]{s} I
   \end{pmatrix} d W.
\]

\subsection{Derivation of the hp-dependent kinetic Fokker-Planck equation}
\label{subsec: kinetic-FP}

First, we derive the corresponding It\^o's formula for the hp-dependent SDE~\eqref{eqn: lr-momentum-sde-standard} as

\begin{lem}[It\^o's lemma]
\label{lem: ito-lemma}
\; For any $f \in C^{\infty}(\mathbb{R}^d)$ and $u \in C^{\infty}\left( [0, +\infty) \times \mathbb{R}^d \times \mathbb{R}^d \right)$, let $(X, V)$ be the solution to the hp-dependent SDE~\eqref{eqn: lr-momentum-sde-standard}. Then, we have
\begin{align}
d u(t, X, V) &= \left( \frac{\partial u}{\partial t} + V \cdot \nabla_x u  - \nabla_x f  \cdot \nabla_v u - 2\sqrt{\mu} V \cdot \nabla_v u   + \frac{\sqrt{s}}{2}\Delta_v u\right) d t  \nonumber \\
                  & \mathrel{\phantom{=}} + \sqrt[4]{s} \left( \sum_{i=1}^{d} \frac{\partial u}{\partial v_i} \right) d W. \label{eqn: ito-formula}
\end{align}

\end{lem}

Let $g \in C^{\infty}\left( \mathbb{R}^d \times \mathbb{R}^d \right)$. Then, for any $\tau < t$, we assume 
\begin{equation}
\label{eqn: backward-variable}
u(\tau, x, v) = \mathbb{E}[g(X(t), V(t))  | X(\tau) = x, V(\tau) = v]. 
\end{equation}
Directly, from~\eqref{eqn: backward-variable}, we can obtain $u(t, x, v) = g(x, v)$. Hence, we have
\begin{equation}
\label{eqn: expectation-equiv}
\mathbb{E} \left[ u(t, X(t), V(t)) - u(\tau, X(\tau), V(\tau)) \big | X(\tau) = x, V(\tau) = v \right] = 0. 
\end{equation}
According to Ito's integral $\mathbb{E}\left[ \int_{\tau}^{t} h dW \right] =0$, then we get the backward Kolmogorov equation for $u(\tau, x, v)$ defined in~\eqref{eqn: backward-variable} as
\begin{equation}
\label{eqn: backward-kolmogorov}
\left\{ \begin{aligned}
         & \frac{\partial u}{ \partial \tau} = - v \cdot \nabla_x u + \nabla_x f \cdot \nabla_v u + 2\sqrt{\mu} v \cdot \nabla_v u - \frac{\sqrt{s}}{2} \Delta_v u \\
                  & u(t, x, v) = g(x, v).
         \end{aligned} \right.
\end{equation}
Now, let $\rho_{s, \alpha}(t, x, v) = \rho_{s, \alpha}(X(t) = x, V(t) = v)$ be the evolving probability density with the initial as $\rho_{s, \alpha}(0, x, v) $. Taking $\tau = 0$, then~\eqref{eqn: expectation-equiv} can be written as
\begin{align*}
\int_{\mathbb{R}^d} \int_{ \mathbb{R}^d} u(t, y ,w) p(t, y, w| 0, x,v) dydw - u(0, x, v) = 0. 
\end{align*}
According to the Chapman-Kolmogorov equation, we can obtain the expectation $\mathbb{E}[u(t, x, v)]$ is a constant, that is, 
\begin{align*}
\mathbb{E} [u(t, x ,v)] & = \int_{\mathbb{R}^d} \int_{ \mathbb{R}^d} u(t, x ,v) \rho_{s, \alpha}(t, x, v) dxdv \\
                                                    & = \int_{\mathbb{R}^d} \int_{ \mathbb{R}^d} u(t, y ,w) \rho_{s, \alpha}(t, y, w) dydw \\
                                                    & = \int_{\mathbb{R}^d} \int_{ \mathbb{R}^d} u(t, y ,w) \left( \int_{\mathbb{R}^d} \int_{ \mathbb{R}^d} p(t, y, w| 0, x,v) \rho_{s, \alpha}(0, x, v) dxdv \right) dydw \\
                                                    & = \int_{\mathbb{R}^d} \int_{ \mathbb{R}^d}  \left( \int_{\mathbb{R}^d} \int_{ \mathbb{R}^d} u(t, y ,w)  p(t, y, w| 0, x,v)  dydw \right)  \rho_{s, \alpha}(0, x, v) dxdv \\
                                                    & = \int_{\mathbb{R}^d} \int_{ \mathbb{R}^d} u(0, x, v)  \rho_{s, \alpha}(0, x, v) dxdv \\
                                                    & =  \mathbb{E}[u(0, x ,v)].
\end{align*}
Hence, by differentiating the expectation $\mathbb{E}[u(t, x, v)]$, we can obtain
\begin{align*}
\int_{\mathbb{R}^d} \int_{ \mathbb{R}^d} u \frac{\partial \rho_{s, \alpha}}{\partial t} dxdv & = - \int_{\mathbb{R}^d} \int_{ \mathbb{R}^d}  \frac{\partial u}{\partial t}  \rho_{s, \alpha} dxdv \\
                                                                                                                         & =  - \int_{\mathbb{R}^d} \int_{ \mathbb{R}^d} (- v \cdot \nabla_x u + \nabla_x f \cdot \nabla_v u + 2\sqrt{\mu} v \cdot \nabla_v u -\frac{\sqrt{s}}{2} \Delta_v u)  \rho_{s, \alpha} dxdv \\
                                                                                                                         & = \int_{\mathbb{R}^d} \int_{ \mathbb{R}^d} u \left[- v \cdot \nabla_x \rho_{s, \alpha} + \nabla_x f \cdot \nabla_v \rho_{s, \alpha} + 2\sqrt{\mu} \nabla_v \cdot (v\rho_{s, \alpha}) + \frac{\sqrt{s}}{2} \Delta_v \rho _{s, \alpha}\right] dxdv.
\end{align*}
Since $u$ is an arbitrary function, we complete the derivation of the hp-dependent kinetic Fokker-Planck equation.

\subsection{The existence and uniqueness of Gibbs invariant distribution}
\label{subsec: exist-unique-gibbs}
%

Recall the equivalent form of the hp-dependent kinetic Fokker-Planck equation~\eqref{eqn: kinetic-equal1}
\[
\frac{\partial h_{s, \alpha} }{\partial t} =  - \mathscr{L}_{s, \alpha} h_{s, \alpha},
\]
where the linear operator $ \mathscr{L}_{s, \alpha}$ is expressed as
\[
\mathscr{L}_{s, \alpha} h_{s, \alpha} = \mathscr{T} + \mathscr{D}_{s, \alpha} = v \cdot \nabla_x - \nabla_xf \cdot \nabla_v + 2\sqrt{\mu} v \cdot \nabla_v - \frac{\sqrt{s}}{2} \Delta_v.
\]
Now, we start to prove the existence and uniqueness of the Gibbs invariant distribution. The Gibbs invariant distribution $\mu_{s, \alpha}$ is a steady solution. Assume there exists another steady solution $\vartheta_{s, \alpha}$, then $h_{s, \alpha} = \vartheta_{s, \alpha} \mu_{s, \alpha}^{-1}$ should satisfy 
\[
\mathscr{L}_{s, \alpha} h_{s, \alpha} = 0.
\]
Taking integration by parts, for the transport operator $\mathscr{T}$, we have
\begin{align*}
\int_{\mathbb{R}^d} \int_{\mathbb{R}^d} h_{s, \alpha} \mu_{s, \alpha}(\mathscr{T} h_{s, \alpha})  dxdv= 0;
\end{align*}
while for the diffusion operator $\mathscr{D}_{s, \alpha}$, we have
\begin{align*}
\int_{\mathbb{R}^d} \int_{\mathbb{R}^d} h_{s, \alpha}\mu_{s, \alpha} (\mathscr{D}_{s, \alpha} h_{s, \alpha})  dxdv & = \int_{\mathbb{R}^d} \int_{\mathbb{R}^d} h_{s, \alpha}\mu_{s, \alpha} \left( -2\sqrt{\mu}v \cdot \nabla_v + \frac{\sqrt{s}}{2} \Delta_v\right)h_{s, \alpha}  dxdv \\
                                                                                                                                                                                    & =  \int_{\mathbb{R}^d} \int_{\mathbb{R}^d}  -2 \sqrt{\mu} h_{s, \alpha}\mu_{s, \alpha} v \cdot \nabla_v h_{s, \alpha} - \frac{\sqrt{s}}{2} \nabla_{v} h_{s, \alpha} \cdot \nabla_v(h_{s, \alpha} \mu_{s, \alpha}) dxdv \\
                                                                                                                                                                                    & = - \frac{\sqrt{s}}{2} \int_{\mathbb{R}^d} \int_{\mathbb{R}^d} \left\| \nabla_v h_{s, \alpha} \right\|^2 \mu_{s, \alpha} dxdv + \left( -2 \sqrt{\mu} + \frac{\sqrt{s}}{\beta}\right) ... \\
                                                                                                                                                                                    & =  \frac{\sqrt{s}}{2} \int_{\mathbb{R}^d} \int_{\mathbb{R}^d} \left\| \nabla_v h_{s, \alpha} \right\|^2 \mu_{s, \alpha} dxdv. 
\end{align*}
Then, some basic calculations tell us that the linear operator $\mathscr{L}_{s, \alpha}$ satisfies 
\begin{align*}
0 & = \int_{\mathbb{R}^d} \int_{\mathbb{R}^d} h_{s, \alpha} \mathscr{L}_{s, \alpha} h_{s, \alpha} \mu_{s, \alpha} dxdv \\
   & = \int_{\mathbb{R}^d} \int_{\mathbb{R}^d} h_{s, \alpha} \mathscr{T} (h_{s, \alpha} \mu_{s, \alpha}) dxdv+  \int_{\mathbb{R}^d} \int_{\mathbb{R}^d} h_{s, \alpha} \mathscr{D}_{s, \alpha} (h_{s, \alpha} \mu_{s, \alpha} dxdv \\
                                                                                                                                                          & = - \frac{\sqrt{s}}{2} \int_{\mathbb{R}^d} \int_{\mathbb{R}^d} \| \nabla_v h_{s, \alpha} \|^2 \mu_{s, \alpha} dxdv \leq 0.
\end{align*}
Hence, we obtain $h_{s, \alpha}(x, v) = g_{s, \alpha}(x)$. Furthermore, according to $\mathscr{L}_{s, \alpha} h_{s, \alpha} = 0$, we have $-v \cdot \nabla_x g_{s, \alpha} =  0$. Because $v$ is arbitrary, we know $g_{s, \alpha} = C$. With both $\vartheta_{s, \alpha}$ and $\mu_{s, \alpha}$ being probability densities, $C = 1$. Hence, the proof of the existence and uniqueness of the Gibbs invariant distribution is complete.

\subsection{Proof of Lemma~\ref{lem: weak-soln-eu}}
\label{subsec: proof_exist-unique-weak}

Recall that Section~\ref{subsec: connect-KFP-operators} shows that the transition probability density $\rho_{s, \alpha}(t, \cdot, \cdot)\in C^1([0, +\infty), L^{2}(\mu_{s, \alpha}^{-1}))$ governed by the hp-dependent kinetic Fokker-Planck equation~\eqref{eqn: kinetic-FP} is equivalent to the function $\psi_{s}(t, \cdot, \cdot)$ in $C^1([0, +\infty), L^2(\mathbb{R}^{d}, \mathbb{R}^{d}))$ governed by the differential equation~\eqref{eqn: FP-Kramers}. Moreover, in Section~\ref{subsec: connect-KFP-operators}, we have shown that the spectrum of the Kramers operator $\mathscr{K}_{s, \alpha}$ satisfies
\[
0 = \zeta_{s, \alpha; 0} < \zeta_{s, \alpha; 1}  \leq \cdots \leq \zeta_{s, \alpha; \ell} \leq \cdots < +\infty.
\]
Since $L^{2}(\mathbb{R}^{d}, \mathbb{R}^{d})$ is a Hilbert space, there exists a standard orthogonal basis corresponding to the spectrum of the Kramers operator $\mathscr{K}_{s, \alpha}$:
\[
\mu_{s, \alpha} = \phi_{s, \alpha; 0},\; \phi_{s,\alpha; 1},\; \ldots,\; \phi_{s,\alpha; \ell}, \; \ldots \in L^{2}(\mathbb{R}^{d}, \mathbb{R}^{d}).
\]
Then, for any initialization $\psi_{s}(0, \cdot, \cdot) \in L^{2}(\mathbb{R}^{d}, \mathbb{R}^{d})$, there exist a family of constants $c_{\ell}$ ($\ell = 1,2, \ldots$) such that 
\[
\psi_{s, \alpha}(0, \cdot, \cdot) = \sqrt{\mu_{s, \alpha}} + \sum_{\ell=1}^{+\infty}c_{\ell} \phi_{s, \alpha; \ell}. 
\]
Thus, the solution to the partial differential equation~\eqref{eqn: FP-Kramers} is 
\[
\psi_{s, \alpha}(t, \cdot, \cdot) = \sqrt{\mu_{s, \alpha}} + \sum_{\ell=1}^{+\infty}c_{\ell}e^{-\zeta_{s, \alpha; \ell}t} \phi_{s, \alpha; \ell}. 
\]
Recognizing the transformation $\psi_{s, \alpha}(t, \cdot, \cdot ) =\rho_{s, \alpha}(t, \cdot, \cdot)/\sqrt{\mu_{s, \alpha}}$, we recover 
\[
\rho_{s, \alpha}(t, \cdot, \cdot) = \mu_{s, \alpha} + \sum_{\ell=1}^{+\infty}c_{\ell}e^{-\zeta_{s, \alpha; \ell}t} \phi_{s, \alpha; \ell} \sqrt{\mu_{s, \alpha}}.
\]
Note that $\zeta_{s, \alpha; \ell}$ is positive for $\ell \geq 1$. Thus, the proof is complete.

%% file: appendix/discrete.tex
\section{Technical Details for~\Cref{sec: main-result}}                                                                                                                          
\label{sec: main-technical}

\subsection{Proof of Proposition~\ref{prop: approx}}
\label{subsec: proof-discrete}


By Lemma~\ref{lem: weak-soln-eu}, let $\rho_{s, \alpha}(t, \cdot, \cdot) \in C^{1}([0, +\infty), L^2(\mu_{s, \alpha}^{-1}))$ denote the unique transition probability density of the solution to the hp-dependent SDE~\eqref{eqn: lr-momentum-sde-standard}. Taking an expectation, we get
\[
\mathbb{E}[X_{s, \alpha}(t)] = \int_{\mathbb{R}^{d}}  \int_{\mathbb{R}^{d}} x \rho_{s, \alpha}(t, x, v) dv d x  .
\]
Hence, the uniqueness has been proved. Using the Cauchy--Schwarz inequality, we obtain:
\begin{align*}
\left\| \mathbb{E}[X_{s, \alpha}(t)] \right\| & \leq  \left\| \int_{\mathbb{R}^{d}} \int_{\mathbb{R}^{d}}  x (\rho_{s}(t, \cdot, \cdot) - \mu_{s, \alpha}) dv d x   \right\| +  \left\| \int_{\mathbb{R}^{d}} \int_{\mathbb{R}^{d}} x  \mu_{s, \alpha} dv d x   \right\| \\
                                                & \leq \left(\int_{\mathbb{R}^d} \int_{\mathbb{R}^{d}}  \|x\|^{2} \mu_{s, \alpha} dvdx \right)^{\frac{1}{2}} \left( \left\| \rho_{s, \alpha}(t, \cdot, \cdot) - \mu_{s, \alpha} \right\|_{L^2(\mu_s^{-1})} + 1 \right)    \\
                                                & \leq  \left(\int_{\mathbb{R}^d} \int_{\mathbb{R}^{d}}  \|x\|^{2} \mu_{s, \alpha} dvdx \right)^{\frac{1}{2}} \left( e^{-\lambda_{s, \alpha}t}\left\| \rho_{s, \alpha}(0, \cdot, \cdot) - \mu_{s, \alpha} \right\|_{L^2(\mu_s^{-1})}+ 1 \right)  \\
                                                & < + \infty,   
 \end{align*}
where the integrability $\int_{\mathbb{R}^d} \int_{\mathbb{R}^{d}}  \|x\|^{2} \mu_{s, \alpha} dvdx  $ is due to the fact that the objective $f$ satisfies the Villani conditions. The existence of a global solution to the hp-dependent SDE~\eqref{eqn: lr-momentum-sde-standard} is thus esta- -blished.

For the strong convergence, the~hp-dependent SDE~\eqref{eqn: lr-momentum-sde-standard} corresponds to the Milstein scheme in numerical methods. The original result is obtained by Milstein~\cite{mil1975approximate} and Talay~\cite{talay1982analyse, pardoux1985approximation}, independently. We refer the readers to~\cite[Theorem 10.3.5 and Theorem 10.6.3]{kloeden1992approximation}, which studies numerical schemes for the stochastic differential equation. For the weak convergence, we can obtain numerical errors by using both the Euler-Maruyama scheme and the Milstein scheme. The original result is obtained by Milstein~\cite{mil1986weak} and Talay~\cite{pardoux1985approximation, talay1984efficient} independently and~\cite[Theorem 14.5.2]{kloeden1992approximation} is
also a well-known reference. Furthermore, a more accurate estimate of $B(T)$ is shown in~\cite{bally1996law}.
The original proofs in the references above only assume finite smoothness, such as $C^{6}(\mathbb{R}^d)$ for the objective function. 


%% file: appendix/Fokker-Planck.tex
\section{Technical Details for~\Cref{sec: qualitative}}                                                                                                                          
\label{sec: technical-prelim}

\subsection{Proof of Lemma~\ref{lem: basic-operator}}
\label{subsec: proof-lemma41}
Here, we show the basic calculations in detail for the readers of all fields. Since the calculations are basic operations, if the readers are familiar with them, you can ignore them. 
\begin{enumerate}[label=\text{(\roman*)}]
\item For any $g_1, g_2 \in L^2(\mu_{s, \alpha})$, with the definition of conjugate linear operators, we have 
\begin{align*}
\langle \mathscr{A}_{s, \alpha}^\star g_1, g_2 \rangle_{1} = \langle  g_1, \mathscr{A}_{s, \alpha} g_2 \rangle_{1} & = \sqrt[4]{\frac{s}{4}} \int_{\mathbb{R}^d}  \int_{\mathbb{R}^d}  \left( g_1 \cdot \nabla_v g_2 \right) \mu_{s, \alpha} \dd v\dd x \\
                                                                                                 & = - \sqrt[4]{\frac{s}{4}}  \int_{\mathbb{R}^d \times \mathbb{R}^d} \left[\nabla_v \cdot (g_1 \mu_{s, \alpha}) \right]g_2  \dd v\dd x     \\
                                                                                                 & = - \sqrt[4]{\frac{s}{4}}  \int_{\mathbb{R}^d \times \mathbb{R}^d} \left( \nabla_v \cdot g_1 - \frac{2 v}{\beta}g_1 \right) g_2 \mu_{s, \alpha}   \dd v\dd x  \\
                                                                                                 & = \left\langle \sqrt[4]{\frac{s}{4}} \left( - \nabla_v + \frac{2v}{\beta} \right) g_1, g_2   \right\rangle_{1}                                                                                            
\end{align*}
and 
\begin{align*}
\langle \mathscr{C}_{s, \alpha}^\star g_1, g_2 \rangle_{1}  = \langle  g_1, \mathscr{C} g_2 \rangle_{1}  & = \sqrt[4]{\frac{s}{4}} \int_{\mathbb{R}^d}  \int_{\mathbb{R}^d }   \left( g_1 \cdot \nabla_x g_2 \right) \mu_{s, \alpha}  \dd v\dd x \\
                                                                                                 & = - \sqrt[4]{\frac{s}{4}} \int_{\mathbb{R}^d}\int_{\mathbb{R}^d} \left[\nabla_x \cdot (g_1 \mu_{s, \alpha}) \right]g_2  \dd v\dd x     \\
                                                                                                 & = - \sqrt[4]{\frac{s}{4}} \int_{\mathbb{R}^d \times \mathbb{R}^d} \left( \nabla_x \cdot g_1 - \frac{2 \nabla_x f}{\beta}g_1 \right) g_2 \mu_{s, \alpha}   \dd v\dd x      \\
                                                                                                 & =  \left\langle \sqrt[4]{\frac{s}{4}} \left( - \nabla_x + \frac{2\nabla_xf}{\beta} \right) g_1, g_2   \right\rangle_{1}.                                                                                         
\end{align*}
Hence, we obtain the diffusion operator as $\mathscr{D}_{s, \alpha} = \mathscr{A}_{s, \alpha}^\star \mathscr{A}_{s, \alpha}$.

\item Since both $\mathscr{A}_{s, \alpha}$ and $\mathscr{A}_{s, \alpha}^\star$ are only the linear operators about $v$ and the linear operator $\mathscr{C}_{s, \alpha}$ about $x$, both $\mathscr{A}_{s, \alpha}$ and $\mathscr{A}_{s, \alpha}^\star$ commutes with $\mathscr{C}_{s, \alpha}$, that is,
\[
[\mathscr{A}_{s, \alpha}, \mathscr{C}_{s, \alpha}] = [\mathscr{A}_{s, \alpha}^\star, \mathscr{C}_{s, \alpha}] = 0.
\]
For the commutator between $\mathscr{A}$ and $\mathscr{A}^\star$, we have
 \begin{align*}
         [\mathscr{A}, \mathscr{A}^\star] & = \frac{\sqrt{s}}{2} \left[ \nabla_v, - \nabla_v + \frac{2v}{\beta} \right] \\
                                                            & =  \frac{\sqrt{s}}{2} \left[ \nabla_v  \left( - \nabla_v + \frac{2v}{\beta} \right) - \left( - \nabla_v + \frac{2v}{\beta} \right)   \nabla_v \right] \\
                                                            & = 2\sqrt{\mu} \mathbf{I}_{d \times d}.
\end{align*} 
         
\item Taking the simple equation $\left(   v \cdot \nabla_x - \nabla_{x} f \cdot \nabla_{v} \right)\mu_{s, \alpha} = 0$, for any $g_1, g_2 \in L^2(\mu_{s, \alpha})$,we have
         \begin{align*}
         \left\langle \mathscr{T}g_1, g_2 \right\rangle_{1} & = \int_{\mathbb{R}^d} \int_{\mathbb{R}^d} \left(   v \cdot \nabla_x - \nabla_{x} f \cdot \nabla_{v} \right)g_1 g_2 \mu_{s, \alpha}\dd v \dd x \\
                                                                                                                    & = - \int_{\mathbb{R}^d} \int_{\mathbb{R}^d} \left(   v \cdot \nabla_x - \nabla_{x} f \cdot \nabla_{v} \right)(g_2 \mu_{s, \alpha}) g_1\dd v \dd x \\
                                                                                                                    & = - \int_{\mathbb{R}^d} \int_{\mathbb{R}^d} \left(   v \cdot \nabla_x - \nabla_{x} f \cdot \nabla_{v} \right)g_2 g_1 \mu_{s, \alpha}\dd v \dd x \\
                                                                                                                    & = - \left\langle g_1, \mathscr{T}g_2 \right\rangle_{1}.
         \end{align*}
The commutator between $\mathscr{A}_{s, \alpha} $ and $\mathscr{T}$ is
\begin{align*}
[\mathscr{A}_{s, \alpha}, \mathscr{T}] & = \mathscr{A}_{s, \alpha} \mathscr{T} - \mathscr{T} \mathscr{A}_{s, \alpha}  \\
                                                 & = \sqrt[4]{\frac{s}{4}}\bigg[ \nabla_v  \left(   v \cdot \nabla_x - \nabla_{x} f \cdot \nabla_{v} \right) - \left(   v \cdot \nabla_x -\nabla_{x} f \cdot \nabla_{v} \right) \nabla_v \bigg] \\
                                                 & =  \sqrt[4]{\frac{s}{4}} \nabla_x = \mathscr{C}_{s, \alpha}. 
\end{align*}
Similarly, the commutator between $\mathscr{C}_{s, \alpha} $ and $\mathscr{T}$ is
         \begin{align*}
         [\mathscr{C}_{s, \alpha}, \mathscr{T}] &  = \mathscr{C}_{s, \alpha} \mathscr{T} - \mathscr{T} \mathscr{C}_{s, \alpha} \\
                                                   & = \sqrt[4]{\frac{s}{4}} \bigg[ \nabla_x \left( v \cdot \nabla_x - \nabla_{x} f \cdot \nabla_{v}\right) - \left( v \cdot \nabla_x - \nabla_{x} f \cdot \nabla_{v}\right) \nabla_x  \bigg] \\
                                                   & = - \sqrt[4]{\frac{s}{4}} \nabla_x^2 f \cdot \nabla_v.
         \end{align*}

%
\end{enumerate}

\subsection{Proof of Lemma~\ref{lem: l2-l1}}
\label{subsec: proof-l2-l1}

For any $g \in L^{2}(\mu_{s, \alpha}^{-1})$, we can write down it as 
\[
\int_{\mathbb{R}^d} \int_{\mathbb{R}^d} |g(v, x)|^2 \mu_{s, \alpha}^{-1}dvdx < +\infty.
\]
Then, we can take the following simple calculation as 
\begin{align*}
\int_{\mathbb{R}^d} \int_{\mathbb{R}^d} | g(x, v) | dvdx & = \int_{\mathbb{R}^d} \int_{\mathbb{R}^d} | g(x, v) |  \mu_{s, \alpha}^{-\frac{1}{2}}   \mu_{s, \alpha}^{\frac{1}{2}}  dvdx  \\
                                                                                       & \leq  \left( \int_{\mathbb{R}^d} \int_{\mathbb{R}^d} | g(x, v) |^{2}  \mu_{s, \alpha}^{-1}  dvdx \right)^{\frac{1}{2}} \left( \int_{\mathbb{R}^d} \int_{\mathbb{R}^d}  \mu_{s, \alpha}  dvdx  \right)^{\frac{1}{2}} \\
                                                                                       & =   \left( \int_{\mathbb{R}^d} \int_{\mathbb{R}^d} | g(x, v) |^{2}  \mu_{s, \alpha}^{-1}  dvdx \right)^{\frac{1}{2}} < +\infty.
\end{align*}
This completes the proof. 
\subsection{Technical Details in Section~\ref{subsubsec: h-1-norm}}
\label{subsec: technical-details-h1}

Here, we compute the derivative in detail as
\begin{align*}
\frac{1}{2} \frac{d}{d t} \left\langle h_{s, \alpha}, h_{s, \alpha}  \right\rangle_{H^{1}(\mu_{s, \alpha})} 
                                                                                                                                               &= - \underbrace{\left\langle \mathscr{L}_{s, \alpha}h_{s, \alpha}, h_{s, \alpha} \right\rangle_{L^{2}(\mu_{s, \alpha})}}_{\mathbf{I}} - \underbrace{\left\langle \mathscr{A}_{s, \alpha}  \mathscr{L}_{s, \alpha}h_{s, \alpha}, \mathscr{A}_{s, \alpha} h_{s, \alpha}  \right\rangle_{L^{2}(\mu_{s, \alpha})}}_{\mathbf{II}} \\ &\mathrel{\phantom{=}} - \underbrace{ \left\langle \mathscr{C}_{s, \alpha}  \mathscr{L}_{s, \alpha} h_{s, \alpha}, \mathscr{C}_{s, \alpha} h_{s, \alpha}  \right\rangle_{L^{2}(\mu_{s, \alpha})}}_{\mathbf{III}}. 
\end{align*}

\begin{enumerate}[label=\textbf{(\arabic*)}]
\item For the term $\mathbf{I}$, we have 
        \[
        \mathbf{I} = \left\langle \mathscr{L}_{s, \alpha}h_{s, \alpha}, h_{s, \alpha}  \right\rangle_{L^{2}(\mu_{s, \alpha})} =  \left\langle \mathscr{A}_{s, \alpha}h_{s, \alpha}, \mathscr{A}_{s, \alpha}h_{s, \alpha}  \right\rangle_{L^{2}(\mu_{s, \alpha})}  = \left\| \mathscr{A}_{s, \alpha} h_{s, \alpha} \right\|^2_{L^{2}(\mu_{s, \alpha})} .
        \] 

\item For the term $\mathbf{II}$, we have         
         \begin{align*}
         \mathbf{II}  = \underbrace{\left\langle \mathscr{A}_{s, \alpha} \mathscr{A}^\star_{s, \alpha} \mathscr{A}_{s, \alpha} h_{s, \alpha}, \mathscr{A}_{s, \alpha} h_{s, \alpha}  \right\rangle_{L^{2}(\mu_{s, \alpha})} }_{\textbf{II}_a} +  \underbrace{\left\langle \mathscr{A}_{s, \alpha} \mathscr{T} h_{s, \alpha}, \mathscr{A}_{s, \alpha} h_{s, \alpha}  \right\rangle_{L^{2}(\mu_{s, \alpha})} }_{\textbf{II}_b} 
         \end{align*}
         \begin{itemize}
         \item For the term $\mathbf{II}_a$, we have
         \begin{align*}
          \mathbf{II}_a & = \left\langle \mathscr{A}_{s, \alpha} \mathscr{A}^\star_{s, \alpha} \mathscr{A}_{s, \alpha}   h_{s, \alpha}, \mathscr{A}_{s, \alpha} h_{s, \alpha} \right\rangle_{L^{2}(\mu_{s, \alpha})} \\
                               & =  \left\langle \mathscr{A}^\star_{s, \alpha} \mathscr{A}_{s, \alpha} \mathscr{A}_{s, \alpha} h_{s, \alpha}, \mathscr{A} _{s, \alpha}h_{s, \alpha}  \right\rangle_{L^{2}(\mu_{s, \alpha})} + \left\langle \left[\mathscr{A}_{s, \alpha}, \mathscr{A}_{s, \alpha}^\star \right]\mathscr{A}_{s, \alpha}  h_{s, \alpha}, \mathscr{A}_{s, \alpha} h_{s, \alpha}  \right\rangle_{L^{2}(\mu_{s, \alpha})} \\
                               & =  \left\langle\mathscr{A}^2_{s, \alpha} h_{s, \alpha}, \mathscr{A}^2_{s, \alpha} h_{s, \alpha}  \right\rangle_{L^{2}(\mu_{s, \alpha})} + \left\langle \left[\mathscr{A}_{s, \alpha}, \mathscr{A}^\star_{s, \alpha} \right]\mathscr{A}_{s, \alpha}   h_{s, \alpha}, \mathscr{A}_{s, \alpha} h_{s, \alpha}  \right\rangle_{L^{2}(\mu_{s, \alpha})} \\
                               & =  \left\langle\mathscr{A}^2_{s, \alpha} h_{s, \alpha}, \mathscr{A}^2_{s, \alpha} h_{s, \alpha}  \right\rangle_{L^{2}(\mu_{s, \alpha})} + 2\sqrt{\mu}\left\langle \mathscr{A}_{s, \alpha} h_{s, \alpha}, \mathscr{A}_{s, \alpha} h_{s, \alpha}  \right\rangle_{L^{2}(\mu_{s, \alpha})}    \\
                               & = \| \mathscr{A}_{s, \alpha}^2 h_{s, \alpha} \|^2_{L^{2}(\mu_{s, \alpha})}  + 2\sqrt{\mu}\| \mathscr{A}_{s, \alpha} h_{s, \alpha} \|^2_{L^{2}(\mu_{s, \alpha})} 
         \end{align*} 
         
         \item For the term $\mathbf{II}_b$, we have    
          \begin{align*}
          \mathbf{II}_b & =  \left\langle \mathscr{A}_{s, \alpha} \mathscr{T} h_{s, \alpha}, \mathscr{A}_{s, \alpha} h_{s, \alpha}  \right\rangle_{L^{2}(\mu_{s, \alpha})} \\ & = \left\langle \mathscr{T} \mathscr{A}_{s, \alpha}  h_{s, \alpha}, \mathscr{A}_{s, \alpha} h_{s, \alpha}  \right\rangle_{L^{2}(\mu_{s, \alpha})}  + \left\langle [\mathscr{A}_{s, \alpha}, \mathscr{T}] h_{s, \alpha}, \mathscr{A}_{s, \alpha} h_{s, \alpha}  \right\rangle_{L^{2}(\mu_{s, \alpha})} \\ & =  \left\langle \mathscr{C}_{s, \alpha} h_{s, \alpha}, \mathscr{A}_{s, \alpha} h_{s, \alpha}  \right\rangle_{L^{2}(\mu_{s, \alpha})} 
         \end{align*}    
         \end{itemize}
                           
         \item For the term $\mathbf{III}$, we have         
         \begin{align*}
         \mathbf{III} & = \underbrace{\left\langle \mathscr{C}_{s, \alpha} \mathscr{A}^\star_{s, \alpha} \mathscr{A}_{s, \alpha} h_{s, \alpha}, \mathscr{C}_{s, \alpha} h_{s, \alpha}  \right\rangle_{L^{2}(\mu_{s, \alpha})} }_{\textbf{III}_a} +  \underbrace{\left\langle \mathscr{C}_{s, \alpha} \mathscr{T} h_{s, \alpha}, \mathscr{C}_{s, \alpha} h_{s, \alpha} \right\rangle_{L^{2}(\mu_{s, \alpha})} }_{\textbf{III}_b} 
         \end{align*}    
          \begin{itemize}
          \item[(1)] For the term $\mathbf{III}_a$, we have
                         \begin{align*}                         
                         \mathbf{III}_a & = \left\langle \mathscr{C}_{s, \alpha} \mathscr{A}^\star_{s, \alpha} \mathscr{A}_{s, \alpha} h_{s, \alpha}, \mathscr{C}_{s, \alpha} h_{s, \alpha}  \right\rangle_{L^{2}(\mu_{s, \alpha})} =  \left\langle  \mathscr{A}^\star_{s, \alpha} \mathscr{C}_{s, \alpha} \mathscr{A}_{s, \alpha}   h_{s, \alpha}, \mathscr{C}_{s, \alpha} h_{s, \alpha}  \right\rangle_{L^{2}(\mu_{s, \alpha})} \\
                                               & = \left\langle \mathscr{C}_{s, \alpha} \mathscr{A}_{s, \alpha} h_{s, \alpha},  \mathscr{A}_{s, \alpha} \mathscr{C}_{s, \alpha} h_{s, \alpha}  \right\rangle_{L^{2}(\mu_{s, \alpha})} =  \left\langle \mathscr{A}_{s, \alpha} \mathscr{C}_{s, \alpha} h_{s, \alpha},  \mathscr{A}_{s, \alpha} \mathscr{C}_{s, \alpha} h_{s, \alpha} \right\rangle_{L^{2}(\mu_{s, \alpha})} \\ &= \| \mathscr{A}_{s, \alpha} \mathscr{C}_{s, \alpha}   h_{s, \alpha} \|^2_{L^{2}(\mu_{s, \alpha})}
                         \end{align*}
          
          \item[(2)] For the term $\mathbf{III}_b$, we have 
                         \begin{align*}                         
                         \mathbf{III}_b & = \left\langle \mathscr{C}_{s, \alpha} \mathscr{T} h_{s, \alpha}, \mathscr{C}_{s, \alpha} h_{s, \alpha}  \right\rangle_{L^{2}(\mu_{s, \alpha})} \\ &= \left\langle \mathscr{T} \mathscr{C}_{s, \alpha}  h_{s, \alpha}, \mathscr{C}_{s, \alpha} h_{s, \alpha} \right\rangle_{L^{2}(\mu_{s, \alpha})} + \left\langle \left[\mathscr{C}_{s, \alpha}, \mathscr{T}\right] h_{s, \alpha}, \mathscr{C}_{s, \alpha} h_{s, \alpha}  \right\rangle_{L^{2}(\mu_{s, \alpha})} \\
                         & = \left\langle \left[\mathscr{C}_{s, \alpha}, \mathscr{T}\right] h_{s, \alpha}, \mathscr{C}_{s, \alpha} h_{s, \alpha}  \right\rangle_{L^{2}(\mu_{s, \alpha})} \\
                         & = - \left\langle \nabla^2_x f \cdot \mathscr{A}_{s, \alpha} h_{s, \alpha}, \mathscr{C}_{s, \alpha} h_{s, \alpha}  \right\rangle_{L^{2}(\mu_{s, \alpha})}
                         \end{align*}                   
          \end{itemize}
\end{enumerate}

%
%

\subsection{Proof of Lemma~\ref{lem: relative-bound}}
\label{subsec: proof-lemma45}

\hfill
\begin{itemize}
\item[(i)] By a density argument, we may assume that $g$ is smooth and decays fast enough at infinity. Then, taking the identity $\nabla_x \mu_{s, \alpha} = - \left( 2 \nabla f/ \beta(s, \alpha) \right) \mu_{s, \alpha}$ and by integration by parts, we have
\begin{align*}
\int_{\mathbb{R}^d} \int_{\mathbb{R}^d} \| \nabla_x f \|^2 g^2 \mu_{s, \alpha} dx dv& = - \frac{\beta(s, \alpha)}{2} \int_{\mathbb{R}^d}\int_{\mathbb{R}^d} g^2 \nabla_x f \cdot \nabla_x \mu_{s, \alpha} dx dv \\
                                                                            & =   \frac{\beta(s, \alpha)}{2} \int_{\mathbb{R}^d} \int_{\mathbb{R}^d} \nabla_x \cdot \left( g^2 \nabla_x f \right) \mu_{s, \alpha} dx dv \\
                                                                            & =   \frac{\beta(s, \alpha)}{2} \int_{\mathbb{R}^d} \int_{\mathbb{R}^d} g^2 (\Delta_x f)  \mu_{s, \alpha} dx dv + \beta(s, \alpha) \int_{\mathbb{R}^d} \int_{\mathbb{R}^d} g \left( \nabla_x g \cdot \nabla_x f \right) \mu_{s, \alpha} dxdv
\end{align*}
By Cauchy-Schwarz inequality
\begin{align*}
\int_{\mathbb{R}^d} \int_{\mathbb{R}^d} \| \nabla_x f \|^2 g^2 \mu_{s, \alpha}dxdv \leq & \frac{\beta(s, \alpha)}{2} \sqrt{ \int_{\mathbb{R}^d} \int_{\mathbb{R}^d} g^2 (\Delta_x f)^2 \mu_{s, \alpha} dxdv } \sqrt{ \int_{\mathbb{R}^d}\int_{\mathbb{R}^d} g^2  \mu_{s, \alpha} dxdv } \\
                                                                                  & +  \beta(s, \alpha) \sqrt{ \int_{\mathbb{R}^d} \int_{\mathbb{R}^d} \| \nabla_x f \|^2 g^2 \mu_{s, \alpha} dxdv } \sqrt{ \int_{\mathbb{R}^d} \int_{\mathbb{R}^d} \| \nabla_x g \|^2 \mu_{s, \alpha} dxdv}
\end{align*}
With the Villani Condition-(II), $\| \nabla^2_x f \|_2 \leq C (1 + \| \nabla f \|)$, we have
\[
(\Delta f)^2 \leq d^2 \| \nabla^2 f \|^2_2 \leq d^2 C^2 \left( 1 + \| \nabla f \| \right)^2 \leq 4d^2C^2 \left( 1 + \| \nabla f \|^2 \right).
\]
Hence, we can obtain 
\begin{align*}
       & \int_{\mathbb{R}^d} \int_{\mathbb{R}^d} \| \nabla_x f \|^2 g^2\mu_{s, \alpha}dx dv \\
\leq  & dC\beta(s, \alpha) \sqrt{  \int_{\mathbb{R}^d} \int_{\mathbb{R}^d} g^2  \mu_{s, \alpha} dxdv +  \int_{\mathbb{R}^d} \int_{\mathbb{R}^d} \| \nabla_x f \|^2 g^2  \mu_{s, \alpha} dxdv } \sqrt{ \int_{\mathbb{R}^d} \int_{\mathbb{R}^d} g^2  \mu_{s, \alpha} dxdv } \\
                                                                             & +  \beta(s, \alpha) \sqrt{ \int_{\mathbb{R}^d} \int_{\mathbb{R}^d} \| \nabla_x f \|^2 g^2 \mu_{s, \alpha} dxdv } \sqrt{ \int_{\mathbb{R}^d} \int_{\mathbb{R}^d} \| \nabla_x g \|^2 \mu_{s, \alpha} dxdv} \\
\leq  &dC\beta(s, \alpha) \left(\int_{\mathbb{R}^d}\int_{\mathbb{R}^d} g^2  \mu_{s, \alpha} dx dv + \sqrt{ \int_{\mathbb{R}^d} \int_{\mathbb{R}^d} \| \nabla_x f \|^2 g^2  \mu_{s, \alpha} dxdv } \sqrt{ \int_{\mathbb{R}^d} \int_{\mathbb{R}^d} g^2  \mu_{s, \alpha} dxdv } \right) \\
                                                                             & +  \beta(s, \alpha) \sqrt{ \int_{\mathbb{R}^d} \int_{\mathbb{R}^d} \| \nabla_x f \|^2 g^2 \mu_{s, \alpha} dxdv } \sqrt{ \int_{\mathbb{R}^d} \int_{\mathbb{R}^d} \| \nabla_x g \|^2 \mu_{s, \alpha} dxdv} \\
                                                                              \leq & dC\beta(s, \alpha) \int_{\mathbb{R}^d}  \int_{\mathbb{R}^d} g^2  \mu_{s, \alpha} dxdv +  \frac{1}{4} \int_{\mathbb{R}^d}  \int_{\mathbb{R}^d} \| \nabla_x f \|^2 g^2 \mu_{s, \alpha}dxdv + d^2C^2 \beta^2(s, \alpha) \int_{\mathbb{R}^d} \int_{\mathbb{R}^d}g^2 \mu_{s, \alpha} dxdv  \\
                                                                              & +  \frac{1}{4} \int_{\mathbb{R}^d}\int_{\mathbb{R}^d}  \| \nabla_x f \|^2 g^2 \mu_{s, \alpha} dxdv +  \beta^2(s,\alpha) \int_{\mathbb{R}^d}  \int_{\mathbb{R}^d} \| \nabla_x g \|^2 \mu_{s, \alpha} dx dv
\end{align*}
where the last inequality follows Cauchy-Schwartz inequality. Therefore, we have
\begin{align*}
\frac{1}{2} \int_{\mathbb{R}^d} \int_{\mathbb{R}^d} \| \nabla_x f \|^2 g^2 \mu_{s, \alpha} dxdv \leq   \big(  dC\beta(s, \alpha) +  d^2 & C^2 \beta^2(s, \alpha) \big) \int_{\mathbb{R}^d} \int_{\mathbb{R}^d} g^2  \mu_{s, \alpha} dxdv \\ &+ \;\beta^2(s,\alpha) \int_{\mathbb{R}^d}  \int_{\mathbb{R}^d} \| \nabla_x g \|^2 \mu_{s, \alpha} dx dv.
\end{align*}
Multiplied by two, taking $\kappa_1(s, \alpha) = \max\{ 2(  dC\beta(s, \alpha) +  d^2  C^2 \beta^2(s, \alpha) ), \; 2\beta^2(s,\alpha) \}$, we have
\[
\| (\nabla_x f) g \|_{L^2(\mu_{s, \alpha})}^{2} \leq  \kappa_1(s, \alpha) \left( \| g \|_{L^2(\mu_{s, \alpha})}^{2} + \| \nabla_x g \|_{L^2(\mu_{s, \alpha})}^{2} \right).\]

\item[(ii)] Similarly, with the Villani Condition-(II), $\| \nabla^2 f \|^{2}_{2} \leq 2C^2 \left( 1 + \| \nabla f \|^2 \right)$, we have
\begin{align*}
\left\| \|\nabla_x^2 f\|_2 g \right\|_{L^{2}(\mu_{s, \alpha})}^{2} & = \int_{\mathbb{R}^d} \int_{\mathbb{R}^d}  \| \nabla_x^2 f \|_2^{2}  \cdot \| g \|^2 \mu_{s, \alpha} dx dv \\
& \leq 2C^2 \left( \int_{\mathbb{R}^d} \int_{\mathbb{R}^d} \| g \|^2 \mu_{s, \alpha} dxdv  + \int_{\mathbb{R}^d} \int_{\mathbb{R}^d}  \| \nabla_x f \|^{2} \| g \|^2 \mu_{s, \alpha} dx dv\right). 
\end{align*}
With inequality (i), we can obtain directly

\begin{align*}
\left\| \|\nabla_x^2 f\|_2 g \right\|_{L^{2}(\mu_{s, \alpha})}^{2} 
                                                                            &\leq \;  2C^2( 1+ \kappa_1(s, \alpha)) \left(  \int_{\mathbb{R}^d} \int_{\mathbb{R}^d}  \| \nabla_x g \|^{2} \mu_{s, \alpha} dxdv + \int_{\mathbb{R}^d}   \int_{\mathbb{R}^d}  \| g \|^2 \mu_{s, \alpha} dx dv  \right) \\
                                                                            & = \; 2C^2( 1+ \kappa_1(s, \alpha)) \left( \| g \|_{L^2(\mu_{s, \alpha})}^{2} +  \| \nabla_x g \|_{L^2(\mu_{s, \alpha})}^{2}  \right)   
\end{align*}
Taking $\kappa_2(s, \alpha) = 2C^2( 1+ \kappa_1(s, \alpha))$, we complete the proof. 
\end{itemize}

\subsection{Technical Details in Section~\ref{subsubsec: mixed-term}}
\label{subsec: technical-detail-mixed}
Here, we compute the derivative of the mixed term in detail as
\begin{align*}
\frac{d}{d t} \langle \mathscr{A}_{s, \alpha} h_{s, \alpha}, \;&\mathscr{C}_{s, \alpha} h_{s, \alpha} \rangle_{L^2(\mu_{s, \alpha})}   \\
& = \underbrace{ \left\langle \mathscr{A}_{s, \alpha}  \mathscr{L}_{s, \alpha}  h_{s, \alpha}, \mathscr{C}_{s, \alpha}   h_{s, \alpha} \right\rangle_{L^2(\mu_{s, \alpha})} +  \left\langle \mathscr{A}_{s, \alpha} h_{s, \alpha}, \mathscr{C}_{s, \alpha}  \mathscr{L}_{s, \alpha}  h_{s, \alpha} \right\rangle_{L^2(\mu_{s, \alpha})} }_{\textbf{IV}} \\
                                                                                                                                   & =   \underbrace{\left\langle \mathscr{A}_{s, \alpha}  \mathscr{A}^\star \mathscr{A}_{s, \alpha}  h_{s, \alpha}, \mathscr{C}_{s, \alpha}   h_{s, \alpha} \right\rangle_{L^2(\mu_{s, \alpha})} +  \left\langle \mathscr{A}_{s, \alpha}   h_{s, \alpha}, \mathscr{C}_{s, \alpha}   \mathscr{A}^\star_{s, \alpha} \mathscr{A}_{s, \alpha}  h_{s, \alpha} \right\rangle_{L^2(\mu_{s, \alpha})} }_{\textbf{IV}_a} \\ & \mathrel{\phantom{=}} + \underbrace{  \left\langle \mathscr{A}_{s, \alpha}  \mathscr{T}  h_{s, \alpha}, \mathscr{C}_{s, \alpha}   h_{s, \alpha}\right\rangle_{L^2(\mu_{s, \alpha})} +  \left\langle \mathscr{A}_{s, \alpha}   h_{s, \alpha}, \mathscr{C}_{s, \alpha}  \mathscr{T}  h_{s, \alpha} \right\rangle_{L^2(\mu_{s, \alpha})} }_{\textbf{IV}_b}
\end{align*}
\begin{itemize}
\item For the term $\textbf{IV}_a$, we have
       \begin{align*}
       \textbf{IV}_a & = \left\langle \mathscr{A}_{s, \alpha}  \mathscr{A}^\star_{s, \alpha} \mathscr{A}_{s, \alpha}  h_{s, \alpha}, \mathscr{C}_{s, \alpha}   h_{s, \alpha} \right\rangle_{L^2(\mu_{s, \alpha})} +  \left\langle \mathscr{A}_{s, \alpha} h_{s, \alpha}, \mathscr{C}_{s, \alpha} \mathscr{A}^\star_{s, \alpha} \mathscr{A}  h_{s, \alpha} \right\rangle_{L^2(\mu_{s, \alpha})} \\
                         & =  \left\langle  \mathscr{A}^\star_{s, \alpha}  \mathscr{A}^2_{s, \alpha}  h_{s, \alpha}, \mathscr{C}_{s, \alpha} h_{s, \alpha} \right\rangle_{L^2(\mu_{s, \alpha})}  +  \left\langle  \left[\mathscr{A}_{s, \alpha}, \mathscr{A}^\star_{s, \alpha} \right]  \mathscr{A}_{s, \alpha}  h_{s, \alpha}, \mathscr{C}_{s, \alpha}   h_{s, \alpha} \right\rangle_{L^2(\mu_{s, \alpha})} \\
                         &\mathrel{\phantom{=}}+  \left\langle \mathscr{A}_{s, \alpha} h_{s, \alpha}, \mathscr{A}^\star_{s, \alpha} \mathscr{C}_{s, \alpha}  \mathscr{A}_{s, \alpha}  h_{s, \alpha} \right\rangle_{L^2(\mu_{s, \alpha})} \\
                         & = 2 \left\langle \mathscr{A}^2_{s, \alpha}  h_{s, \alpha},  \mathscr{A}_{s, \alpha} \mathscr{C}_{s, \alpha} h_{s, \alpha} \right\rangle_{L^2(\mu_{s, \alpha})} + 2\sqrt{\mu} \left\langle \mathscr{A}_{s, \alpha} h_{s, \alpha},  \mathscr{C}_{s, \alpha} h_{s, \alpha} \right\rangle_{L^2(\mu_{s, \alpha})} \\
                         & \geq - 2 \| \mathscr{A}^2_{s, \alpha}  h_{s, \alpha} \|_{L^2(\mu_{s, \alpha})} \| \mathscr{A}_{s, \alpha} \mathscr{C}_{s, \alpha} h_{s, \alpha} \|_{L^2(\mu_{s, \alpha})} - 2\sqrt{\mu} \| \mathscr{A}_{s, \alpha}   h_{s, \alpha} \|_{L^2(\mu_{s, \alpha})} \|   \mathscr{C}_{s, \alpha}  h_{s, \alpha}  \|_{L^2(\mu_{s, \alpha})}
       \end{align*}  

\item For the term $\textbf{IV}_b$, we have      
         \begin{align*}
         \textbf{IV}_b & = \left\langle \mathscr{A}_{s, \alpha}  \mathscr{T}  h_{s, \alpha}, \mathscr{C}_{s, \alpha}   h_{s, \alpha} \right\rangle_{L^2(\mu_{s, \alpha})} +  \left\langle \mathscr{A}_{s, \alpha}   h_{s, \alpha}, \mathscr{C}_{s, \alpha}  \mathscr{T}  h_{s, \alpha} \right\rangle_{L^2(\mu_{s, \alpha})} \\
                           & = \left\langle \mathscr{A}_{s, \alpha}  \mathscr{T}  h_{s, \alpha}, \mathscr{C}_{s, \alpha} h_{s, \alpha} \right\rangle_{L^2(\mu_{s, \alpha})} +  \left\langle \mathscr{A}_{s, \alpha}   h_{s, \alpha}, \mathscr{T} \mathscr{C}_{s, \alpha}   h_{s, \alpha} \right\rangle_{L^2(\mu_{s, \alpha})} +  \left\langle \mathscr{A}_{s, \alpha}   h_{s, \alpha}, \left[\mathscr{C}_{s, \alpha}, \mathscr{T}\right]    h_{s, \alpha} \right\rangle_{L^2(\mu_{s, \alpha})} \\
                           & = \left\langle \mathscr{A}_{s, \alpha} \mathscr{T}  h_{s, \alpha}, \mathscr{C}_{s, \alpha}   h_{s, \alpha} \right\rangle_{L^2(\mu_{s, \alpha})} -  \left\langle \mathscr{T} \mathscr{A}_{s, \alpha}   h_{s, \alpha},  \mathscr{C}_{s, \alpha}   h_{s, \alpha} \right\rangle_{L^2(\mu_{s, \alpha})} -  \left\langle \mathscr{A}_{s, \alpha}   h_{s, \alpha}, \left[\mathscr{T}, \mathscr{C}_{s, \alpha} \right]    h_{s, \alpha} \right\rangle_{L^2(\mu_{s, \alpha})} \\
                            & = \left\langle (\mathscr{A}_{s, \alpha}  \mathscr{T} - \mathscr{T} \mathscr{A}_{s, \alpha}) h_{s, \alpha}, \mathscr{C}_{s, \alpha}   h_{s, \alpha} \right\rangle_{L^2(\mu_{s, \alpha})}  -  \left\langle \mathscr{A}_{s, \alpha}  h_{s, \alpha}, \left[\mathscr{T}, \mathscr{C}_{s, \alpha}\right]    h_{s, \alpha} \right\rangle_{L^2(\mu_{s, \alpha})} \\
                            & \geq \| \mathscr{C}_{s, \alpha}   h_{s, \alpha} \|^2_{L^2(\mu_{s, \alpha})} - \sqrt{\kappa_3(s, \alpha)} \| \mathscr{A}_{s, \alpha}   h_{s, \alpha}\|_{L^2(\mu_{s, \alpha})} \left( \|\mathscr{A}_{s, \alpha} \mathscr{C}_{s, \alpha} h_{s, \alpha}  \|_{L^2(\mu_{s, \alpha})} +  \| \mathscr{A}_{s, \alpha} h_{s, \alpha}  \|_{L^2(\mu_{s, \alpha})}  \right)
         \end{align*} 
\end{itemize}